\documentclass[11pt]{article}

\usepackage[utf8]{inputenc} 
\usepackage[T1]{fontenc}    
\usepackage{lmodern}

\usepackage{hyperref}       
\usepackage{url}            
\usepackage{booktabs}       
\usepackage{amsfonts}       
\usepackage{nicefrac}       
\usepackage{microtype}      
\usepackage{xcolor}         

\usepackage{macros}

\usepackage{tikz}
\usepackage{pgfplots}
\pgfplotsset{compat=newest}

\usepackage{bm}
\usepackage{subcaption}

\usepackage[round,authoryear]{natbib}
\defcitealias{wiki:FatousLemma}{``Reverse Fatou's Lemma'' in ProofWiki (2022)}

\usepackage{csquotes}

\title{\bfseries Fine-Grained Distribution-Dependent \\ Learning Curves \\ \vspace*{-0.5em}}

\author{
   \footnotesize
   \begin{tabular}{c} 
     {\bfseries Olivier Bousquet} \\ \textit{Google, Brain Team} \\ \texttt{obousquet@google.com} \\ \\
     {\bfseries \small Shay Moran} \\ \textit{Department of Mathematics} \\ \textit{Department of Computer Science} \\ \textit{Technion -- Israel Institute of Technology;} \\ \textit{Google Research} \\ \texttt{smoran@technion.ac.il} \\
   \end{tabular}
   \and
   \footnotesize
   \begin{tabular}{c} 
     {\bfseries \small Steve Hanneke} \\ \textit{Computer Science Department} \\ \textit{Purdue University} \\ \texttt{steve.hanneke@gmail.com} \\ \\
     {\bfseries \small Jonathan Shafer} \\ \textit{Computer Science Division} \\ \textit{UC Berkeley} \\ \texttt{shaferjo@berkeley.edu} \\~\\~\\
   \end{tabular}
   \and
   \footnotesize
   \begin{tabular}{c} 
     {\bfseries \small Ilya Tolstikhin} \\ \textit{Google, Brain Team} \\ \texttt{tolstikhin@google.com} \\ \vspace*{-2em}
   \end{tabular} 
}

\date{\scriptsize November 2022 \\ \vspace*{-0.2em}}

\begin{document}

\maketitle

\thispagestyle{empty}


\begin{abstract}
	{\scriptsize 
		Learning curves plot the expected error of a learning algorithm as a function of the number of labeled samples it receives from a target distribution. They are widely used as a measure of an algorithm's performance, but classic PAC learning theory cannot explain their behavior.
		
		As observed by \cite{,DBLP:conf/colt/AntosL96,DBLP:journals/ml/AntosL98}, the classic `No Free Lunch' lower bounds only trace the upper envelope above all learning curves of specific target distributions. For a concept class with VC dimension $d$ the classic bound decays like $d/n$, yet it is possible that the learning curve for \emph{every} specific distribution decays exponentially. In this case, for each $n$ there exists a different `hard' distribution requiring $d/n$ samples. \citeauthor{DBLP:conf/colt/AntosL96} asked which concept classes admit a `strong minimax lower bound' -- a lower bound of $d'/n$ that holds for a fixed distribution for infinitely many $n$.
		
		We solve this problem in a principled manner, by introducing a combinatorial dimension called VCL that characterizes the best $d'$ for which $d'/n$ is a strong minimax lower bound. Our characterization strengthens the lower bounds of \cite*{DBLP:conf/stoc/BousquetHMHY21}, and it refines their theory of learning curves, by showing that for classes with finite VCL the learning rate can be decomposed into a linear component that depends only on the hypothesis class and an exponential component that depends also on the target distribution. As a corollary, we recover the lower bound of \cite{,DBLP:conf/colt/AntosL96,DBLP:journals/ml/AntosL98} for half-spaces in $\mathbb{R}^d$.
		
		Finally, to provide another viewpoint on our work and how it compares to traditional PAC learning bounds, we also present an alternative formulation of our results in a language that is closer to the PAC setting.
		
		\vfill
	}
\end{abstract}

\pagebreak

\clearpage

\pagenumbering{arabic} 

\tableofcontents

\vspace*{\baselineskip}

\section{Introduction}

The most fundamental question in learning theory is arguably ``\emph{what can be learned, and what amount of resources (such as data and computation) is necessary for learning when learning is possible?}''
The classic and definitive mathematical treatment of this question for supervised learning has traditionally been provided by the PAC framework, due to \citet{VapChe68} and \citet{DBLP:journals/cacm/Valiant84}.
However, it has become increasingly clear that the PAC model does not accurately capture the reality of learning; VC bounds are overly pessimistic, and modern machine learning algorithms routinely outperform them.
This is partially because the PAC model constitutes a worst-case analysis over all distributions.
In contrast, machine learning practitioners are typically faced with one (or a few) target distributions, they are interested in optimizing performance only with respect to these specific distributions, and therefore they can vastly outdo the worst-case analysis.

Indeed, \cite{DBLP:conf/colt/AntosL96,DBLP:journals/ml/AntosL98} observed that while the classic PAC bounds decay like $\Omega(d/n)$ for a class of VC dimension $d$, there exist hypothesis classes with arbitrarily large VC dimension that are learnable such that for every realizable distribution the expected loss decays exponentially fast. 

They wrote:

\begin{displayquote}
  ``[I]n some sense, these [VC] lower bounds are not satisfactory. They do not tell us anything about the way the error decreases as the sample size is increased for a given classification problem. These bounds, for each $n$, give information about the maximal error within the class, but not about the behavior of the error for a single fixed [distribution] as the sample size $n$ increases. In other words, the `bad' [distribution], causing the largest error for a learning rule, may be different for each $n$.''\footnote{From \cite{DBLP:conf/colt/AntosL96}, edited for clarity.}
\end{displayquote}

This lead them to study the following question:


\begin{ShadedAlgorithmBox}
  \begin{question}[Strong Minimax Lower Bound]\label{question:antos-lugosi}
    For a VC class, what is the largest $d' \geq 0$ such that for every learning algorithm there exists a realizable distribution for which the expected $0$-$1$ loss is at least $d'/n$ infinitely often?
  \end{question}
\end{ShadedAlgorithmBox}

They were able to answer this question for a number of specific hypothesis classes. Furthermore, they showed that ``it is neither the VC dimension, nor the rate of increase of the shatter coefficients of the class'' that determine the answer. The general case, however, has remained open. 

In this paper we solve \cref{question:antos-lugosi}. We do so in a principled manner, by contributing to the nascent study of distribution-dependent learning curves. We build upon the recent results of \cite*{DBLP:conf/stoc/BousquetHMHY21}, who offered a characterization of these curves.

For each instance, consisting of a hypothesis class and a target distribution, the distribution-dependent learning curve is the expected $0$-$1$ loss of a learning algorithm as a function of the number of i.i.d.\ samples from the distribution (see \cref{section:preliminaries} for formal definitions). 

\begin{figure}[H]
  \centering
  \begin{tikzpicture}
    \begin{axis}[width=0.9\textwidth,height=5cm,grid=minor,xmin=0.7,xmax=10.3,ymax=1.1,
      axis x line=middle,
      axis y line=middle,
      x label style={at={(axis description cs:0.5,-0.1)},anchor=north},
      xlabel={$n$ (Number of i.i.d.\ examples)},
      ylabel={Expected Loss},
      y label style={at={(axis description cs:-.01,.5)},rotate=90,anchor=south},
      yticklabels={,,}, y tick label style={major tick length=0pt},
      xticklabels={,,}, x tick label style={major tick length=0pt}
    ]
      \addplot+[domain=1:10,samples=40,mark=none,color=red!80!white,solid] {0.9*exp(1-0.9*x)};
      \addplot+[domain=1:10,samples=40,mark=none,color=red!70!white,solid] {0.8*exp(1-0.8*x)};
      \addplot+[domain=1:10,samples=40,mark=none,color=red!60!white,solid] {0.7*exp(1-0.7*x)};
      \addplot+[domain=1:10,samples=40,mark=none,color=red!50!white,solid] {0.6*exp(1-0.6*x)};
      \addplot+[domain=1:10,samples=40,mark=none,color=red!40!white,solid] {0.5*exp(1-0.5*x)};
      \addplot+[domain=1:10,samples=40,mark=none,color=red!30!white,solid] {0.4*exp(1-0.4*x)};
      \addplot+[domain=1:10,samples=40,mark=none,color=red!20!white,solid] {0.3*exp(1-0.3*x)};
      \addplot+[domain=1:10,samples=40,mark=none,very thick,color=blue,solid] {1/x};
    \end{axis}
    \draw[color=blue,<-] (9,0.65) to[in=180,out=45] (10,1.5) node[right] {$\sim \frac{1}{n}$};
    \draw[color=red,<-] (2.2,0.7) to[in=0,out=215] (1.75,0.35) node[left] {$\sim e^{-c(\cD)\cdot n}$};
    \end{tikzpicture}
    \caption{Illustration of the difference between distribution-dependent and PAC rates. Each red curve shows exponential decay of the error $\Opt = \bbE_{S \sim \cD^n}[L_\cD^{0\text{-}1}(\hh_S)]$ for a different data distribution $\cD$; but the PAC rate only captures the pointwise supremum of these curves (the blue curve) which decays linearly at best.\\\strut \hfill {\footnotesize Source: \cite{DBLP:conf/stoc/BousquetHMHY21}, adapted with permission.}}
    \label{fig:rates}
  \end{figure}
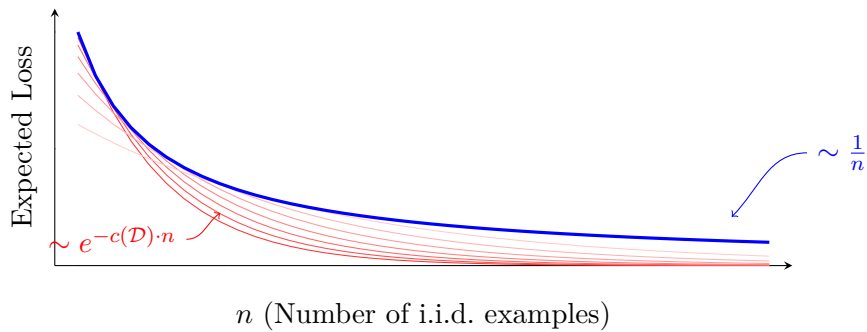
The lay of the land when viewed from the perspective of distribution-dependent learning curves is remarkably structured, and remarkably different from that of the PAC model, as captured by the following crisp trichotomy.

\begin{theorem}
  [\cite{DBLP:conf/stoc/BousquetHMHY21}, Theorem 1.6]
  \label{theorem:trichotomy}
  For every concept class $\cH$ with $|\cH| \geq 3$, exactly one of the following holds:
  \begin{itemize}
    \item $\cH$ is learnable with optimal rate $e^{-n}$.
    \item $\cH$ is learnable with optimal rate $\frac{1}{n}$.
    \item $\cH$ requires arbitrarily slow rates.
  \end{itemize}
\end{theorem}

This differs markedly from PAC learning bounds because, for example, it is possible for a class $\cH$ to have infinite VC dimension but still be learnable with an exponential rate; and ERM algorithms, which are optimal in the PAC setting, can perform arbitrarily worse than the best learning algorithm in the distribution-dependent setting \citep[see Example 2.3 and Example 2.6 respectively in][]{DBLP:conf/stoc/BousquetHMHY21}.
\cite{DBLP:conf/stoc/BousquetHMHY21} also provide a combinatorial characterization (via infinite trees) that determines for each hypothesis class $\cH$ which of the three prongs of the trichotomy it belongs to.

While the trichotomy of \cref{theorem:trichotomy} is an important  characterization, it is far from constituting a complete distribution-dependent theory of supervised learning.
To see this, we recall the definition of \emph{learning at rate $R(n)$} for some function $R: \bbN \rightarrow [0,1]$, as used in the trichotomy. Roughly (see \cref{definition:learning-rates-classic} below), a class $\cH$ is learnable at rate $R$ if there exists a learning algorithm such that for any realizable distribution there exist parameters $C,c \geq 0$ (that depend on the distribution) such that the $0$-$1$ loss of the algorithm after seeing $n$ i.i.d.\ samples from the distribution is at most $C\cdot R(c \cdot n)$.
In other words, each instance, consisting of a hypothesis class and a distribution, determines a pair of parameters $C,c \geq 0$ which together specify the shape of the learning curve.

The characterization of \cref{theorem:trichotomy} explains the general shape of the learning curve (exponential, linear, or arbitrarily slow decay), but it is silent with regard to the parameters $C,c$ that specify its precise shape.
In particular, it is not clear in what manner the class and the distribution `interact' to produce these parameters. This is where the present paper comes in. 

\subsection{Main Results}\label{section:main-results}

Our main contributions are:
\begin{enumerate}
  \item{
    We solve the main question left open by \cite{DBLP:journals/ml/AntosL98,DBLP:conf/colt/AntosL96}. We define a new combinatorial dimension that we call the  Vapnik--Chervonenkis--Littlestone dimension, or VCL, and show that it characterizes the magnitude of the strong minimax lower bound up to universal constants, as follows.
    
    There exist universal constants $\alpha,\beta > 0$ such that for any VC class $\cH$, the dimension $d = \VCL{\cH} \leq \VC{\cH}$, and the number $d'$ defined in question \cref{question:antos-lugosi} satisfies $\alpha \cdot d \leq d' \leq \beta \cdot d$.
  }
  \item{
    More generally, we introduce a more refined characterization of distribution-dependent learning curves. For any class $\cH$, if the dimension $d = \VCL{\cH} \geq 0$ is finite, then the optimal expected loss $\Opt$ can be bounded by
    \begin{equation}\label{eq:general-form-of-learning-curve}
      \frac{\OmegaOf{d}}{n} \leq \Opt \leq \frac{\BigO{d}}{n} + C \cdot e^{-c \cdot n},
    \end{equation}
    where $n \in \bbN$ is the number of i.i.d.\ samples used, and the inequalities hold as follows: for any learning algorithm there exists a distribution such that the lower bound holds for infinitely many values of $n$; the upper bound holds for a learning algorithm that we present, for all distributions and all $n \in \bbN$; the parameters $C,c \geq 0$ depend on $\cH$ and the distribution, and the $\OmegaOf{\cdot}$ and $\BigO{\cdot}$ notations hide universal multiplicative constants that are independent of $\cH$ and of the distribution.
    
    This bound captures both linear rates (when $d > 0$) and exponential rates (when $d = 0$). We call this type of bound the \emph{fine-grained} rate of $\cH$, to distinguish it from the notion of \emph{coarse} rate used in \cref{theorem:trichotomy}. See \cref{{definition:learning-rates-new},theorem:fine-grained} for a formal statement of this result.
  }
  \item{\label{item:main-result-ssl}
    Furthermore, for the hard distribution that satisfies the lower bound in \cref{eq:general-form-of-learning-curve}, the marginal on the domain $\cX$ depends only on the class $\cH$. In particular, in contrast to the lower bounds of \cite{DBLP:conf/stoc/BousquetHMHY21}, the marginal on the domain does not depend on the learning algorithm. Conceptually, this means that in the distribution-dependent learning curve setting, access to unlabeled data is not helpful for learning classes with finite VCL dimension. Namely, semi-supervised learning and supervised learning require the same number of labeled samples.

    We note that this is a non-trivial result, employing a sophisticated application of Fatou's lemma which enables reversing the order of quantifiers, as well as an argument from Ramsey theory.
  }
  \item{
    For any class $\cH$, if $\VCL{\cH} = \infty$ and $\cH$ does not shatter an infinite strong VCL tree (see \cref{definition:strong-vcl-tree,definition:shatter-strong-vcl} below), then $\cH$ has a \emph{strongly distribution-dependent linear rate}. Namely, for every $c \geq 0$ there exists a distribution such that $\Opt \geq \sfrac{c}{n}$ for infinitely many $n \in \bbN$ (see \cref{definition:strongly-distribution-dependent-linear,theorem:fine-grained}).\footnote{In the remaining case where $\VCL{\cH} = \infty$ and $\cH$ has an infinite strong VCL tree, $\cH$ requires arbitrarily slow rates, as shown by \cite{DBLP:conf/stoc/BousquetHMHY21}.}
  }
  \item{
    We offer an equivalent formulation of our results in a language that is closer to the traditional PAC framework. This provides another viewpoint on our work and how it compares to traditional PAC bounds. See \cref{theorem:pac-style}.
  }
  \item{
    As a special case, we recover the lower bound of \cite{antos:98} for half-spaces. We do so by introducing a technique for proving strong lower bounds via a `fractal' argument, which may be useful for other classes as well. (See \cref{theorem:half-spaces,section:half-spaces}.)
  }
\end{enumerate}


\subsection{Benefits of the New Characterization}

Our upper bound in \cref{eq:general-form-of-learning-curve} offers a refinement and reinterpretation of \cref{theorem:trichotomy}:

\begin{itemize}
  \item{
    {\bfseries Upper bound refinement.} For the case of linear rates, \cite{DBLP:conf/stoc/BousquetHMHY21} showed an upper bound of $\frac{c}{n}$, where $c \geq 0$ depends both on the class and on the distribution. In contrast, our expression in the upper bound constitutes a \emph{decomposition} of the error rate into a linear component that depends only on the class, and an exponential component that depends on the class and on the distribution. We view this as a step towards a complete characterization of the optimal distribution-dependent learning rate for supervised learning.
  }
  \item{
    {\bfseries Upper bound reinterpretation.} Whereas \cref{theorem:trichotomy} depicts exponential rates and linear rates as being two entirely different beasts, \cref{eq:general-form-of-learning-curve} presents them in a more unified light, with exponential rates constituting the special case of $\sfrac{d}{n}$ where $d=0$.  
  }
\end{itemize} 

Our lower bound in \cref{eq:general-form-of-learning-curve} offers meaningful improvements over both the previous distribution-dependent lower bound and over the classic `no free lunch' lower bounds from PAC learning, and also constitutes a partial unification of these two results.
\begin{itemize}
  \item{
    {\bfseries Quantitative strengthening of distribution-dependent lower \allowbreak bounds.} For classes with linear learning rates, the best previously known distribution-dependent lower bound that applies to general classes was $\Omega(1/n)$ \citep{DBLP:conf/stoc/BousquetHMHY21}. This applies equally to all classes that have linear rates, and does not distinguish between different degrees of hardness within that broad set of classes. In contrast, we are able to prove a lower bound of $\Opt \geq \OmegaOf{d/n}$, for $d$ that depends only on the class and is tight up to a universal multiplicative constant (independent of the class and of the distribution).
  }
  \item{
    {\bfseries Qualitative \hfill strengthening \hfill of \hfill distribution-dependent \hfill lower \newline bounds.}   
    Classic PAC lower bounds (discussed further in the next bullet) have the following formulation:
    \paragraph*{}\label{quote:stronger-lower-bound-formulation}\vspace*{-2.5em}
    \begin{quotation}
      \noindent \emph{There exists a distribution $\cD_\cX \in \distribution{\cX}$ such that for any learning algorithm $A$ there exists a hard distribution $\cD \in \distribution{\cX \times \{0,1\}}$ such that the marginal distribution of $\cD$ on $\cX$ equals $\cD_\cX$, and the loss of $A$ on distribution $\cD$ is large.}\footnote{The marginal distribution $\cD_\cX$ is simply a uniform distribution over a subset of the domain of cardinality $\VC{\cH}$ that is shattered by $\cH$ in the VC sense.} \hfill $(\star)$ 
    \end{quotation}
    In contrast, the linear lower bound of \cite{DBLP:conf/stoc/BousquetHMHY21} offered a weaker statement: 
    \begin{quotation}
      \noindent \emph{For any learning algorithm $A$ there exists a hard distribution $\cD \in \distribution{\cX \times \{0,1\}}$ such that the loss of $A$ on distribution $\cD$ is large.}
    \end{quotation}
    In particular, this weaker formulation left open the possibility that an algorithm that has access to unlabeled samples (as in the semi-supervised learning setting) could beat the lower bound. We show that that is not the case. We strengthen the lower bound of \cite{DBLP:conf/stoc/BousquetHMHY21}, obtaining the stronger formulation as in (\hyperref[quote:stronger-lower-bound-formulation]{$\star$}).
  }
  \item{
    {\bfseries Qualitative strengthening of PAC lower bounds.} Classic PAC learning theory features `no free lunch' lower bounds \citep[e.g.,][Theorem 5.1]{DBLP:books/daglib/0033642}, which imply the VC lower bound appearing in the fundamental theorem of PAC learning \citep[e.g., ][Theorem 6.8, Item 3]{DBLP:books/daglib/0033642}. These lower bounds leave something to be desired.
    
    To see this, fix a hypothesis class of VC dimension $d$. The VC bound states that for every $\varepsilon > 0$ there exists a hard (worst-case) distribution $\cD_\varepsilon$ such that  
    achieving loss at most $\varepsilon$ with high probability requires at least $\OmegaOf{d/\varepsilon}$ i.i.d.\ samples from $\cD_\varepsilon$. Namely, for a fixed hypothesis class and a sequence of positive values $\varepsilon_1,\varepsilon_2,\dots$ there exists a sequence of \emph{distinct} hard distributions $\cD_1, \cD_2,\dots$ such that each $\cD_i$ is a hard distribution for achieving loss $\varepsilon_i$ --- but it is typically not a hard distribution for other values of $\varepsilon$.

    Clearly, the type of lower bound studied in VC bounds is strictly weaker than the distribution-dependent lower bounds studied in this paper, where there exists a specific hard distribution such that the lower bound holds for infinitely many values of $\varepsilon$. And as we argued above, instance specific lower bounds are a better match to the reality of most machine learning practitioners, who typically face a specific (fixed) unknown distribution, and would like to calculate how many samples are necessary for obtaining loss $\varepsilon_1$, or loss $\varepsilon_2$, or loss $\varepsilon_3$, etc.\ --- all with respect to the \emph{same} fixed unknown distribution. 

    Thus, an interesting open question is ``for which VC classes is it possible to obtain distribution-dependent linear lower bounds of $\Omega(d/n)$?'' (where $d = \VC{\cH}$ and the bound holds for a single distribution for infinitely many $n \in \bbN$). This question, which was studied by \cite{DBLP:conf/colt/AntosL96,DBLP:journals/ml/AntosL98}, is answered by our characterization as follows. Let $\cH$ be a VC class with $0\leq d' = \VCL{\cH} \leq \VC{\cH} = d$. If $d' > 0$ then $\cH$ has a distribution-dependent linear lower bound of $\OmegaOf{\sfrac{d'}{n}}$. Otherwise, if $d' = 0$ then $\cH$ does not have a distribution-dependent linear lower bound; rather, each learning curve decays exponentially and the upper envelope of all the learning curves decays linearly as $\Theta(\sfrac{d}{n})$. In this sense, our results offer a unified perspective of PAC and distribution-dependent lower bounds. 
  }
\end{itemize}


\subsection{Related Works}

\paragraph{Universal Learning.}
Our work explores the distribution-dependent setting, also called the \emph{universal learning} setting, which was recently formalized by \citet*{DBLP:conf/stoc/BousquetHMHY21}.  
	However, it is worthwhile to note that this framework has been studied by earlier works as well.

\citet*{schuurmans:97} revealed the distinction between exponential and linear rates in the universal setting.
	In more detail, \citet*{schuurmans:97} characterized the optimal learning rate for classes $\cH$ that are \emph{concept chains}, 
	meaning that every $h,h' \in \cH$ have either $h \leq h'$ everywhere or $h' \leq h$ everywhere.

\citet*{van-handel:13} studied the uniform convergence property via the universal lens. 
	He characterizes those hypothesis classes $\cH$ satisfying that the empirical losses of \emph{all} hypotheses 
	in the class simultaneously and uniformly converge to the corresponding population losses as the number of examples tends to infinity. 
	The difference with the (more common) distribution-free uniform convergence is that in the universal variant, 
	the rate of the uniform convergence can depend on the source distribution.

An extreme notion of universal learnability is \emph{universal consistency}:
	a learning rule is universally consistent if its expected loss converges to 
	the Bayes optimal risk for every target distribution.
	In other words, such algorithms learn every distribution (but at a distribution-dependent rate).
	The first proof that such learning is possible was provided by \citet*{stone:77} who established the universal consistency
	of several algorithms such as $k$-nearest neighbor predictors, kernel rules, and histogram rules; 
	see \cite*{devroye:96} for a thorough discussion of such results.

\paragraph{No Free Lunch.}
One of the technical contributions in this work is the identification of the VCL dimension 
	as the combinatorial parameter which characterizes when a strong  form of the `no free lunch' theorem holds.
	That is, for which classes is it the case that there exists a single fixed distribution 
	which witnesses the strongest lower bound on the error rate for infinitely many sample sizes $n$. 

The work by \citet*{antos:98} explored this question for VC classes; 
	that is, they asked for which VC classes such a strong `no free lunch' theorem holds. 
	\citet*{antos:98} showed that $d$-dimensional half-spaces satisfy such a strong `no free lunch' theorem 
	by proving a lower bound of $d/n$ on the learning rate. 
	(\citealt*{schuurmans:97} also established such a bound in the $1$-dimensional case.)
	However, a characterization of VC classes with this property remained open; 
	in fact, \citet*{antos:98} explicitly concluded that it is ``neither the VC dimension
	nor the rate of increase of shatter coefficients that determine the asymptotic behavior of the concept class''. 
	Our work resolves this question by showing that the VCL dimension determines this behavior.

\paragraph*{Strong Minimax.} 
A recent work by~\citet*{DBLP:conf/focs/Ben-DavidB20a} studies a similar type of lower bounds for 
	the task of computing boolean functions up to error $\varepsilon$. 
	They introduce a new type of minimax theorem which provides a single hard distribution for arbitrarily small $\varepsilon$.

\section{Preliminaries}\label{section:preliminaries}

\begin{notation}
  $\bbN = \{1,2,3,\dots\}$, i.e., $0 \notin \bbN$. For any $n \in \bbN$, we denote $[n] = \{1,2,3,\dots,n\}$.
\end{notation}

\begin{notation}
  Let $\cX$ be a set. We write $\cX^* = \cup_{t = 0}^\infty \cX^t$ to denote the set of all finite strings or finite vectors with elements from $\cX$. $\cX^*$ includes the empty string, which we denote by $\lambda$.
\end{notation}

\begin{notation}
  For a set $\cX$, we write $\distribution{\cX}$ to denote the set of all distribution with support contained in $\cX$ (with respect to some fixed $\sigma$-algebra). 
\end{notation}

\begin{notation}
  For a (finite or infinite) vector $\bx = (x_1,x_2,\dots)$, we write $\bx_{\leq t}$ to denote the finite prefix $(x_1,x_2,\dots,x_t)$; $\bigl((\bx_t,\by_t)\bigr)_{t \in \bbN}$ denotes an infinite sequence of pairs of vectors, where for each $t$, $(\bx_t,\by_t)$ is a pair of vectors; for a (finite or infinite) sequence of pairs of vectors, we denote a finite prefix of the sequence by $(\bx,\by)_{\leq t} = \bigl((\bx_1,\by_1),(\bx_2,\by_2)\dots,(\bx_t,\by_t)\bigr)$.
\end{notation}

\subsection{Traditional Learning Theory}

\begin{definition}
  Let $\cX$ be a set, and let $\cH \subseteq \{0,1\}^\cX$ be a set of functions. Let $k \in \bbN$, $X = \{x_1,x_2,\dots,x_k\} \subseteq \cX$. We say that \ul{$\cH$ shatters $X$} if for any $y_1,y_2,\dots,y_k \in \{0,1\}$ there exists $h \in \cH$ such that  $h(x_i) = y_i$ for all $i \in [k]$. The \ul{Vapnik--Chervonenkis (VC) dimension of $\cH$}, denoted $\VC{\cH}$, is the largest $d \in \bbN$ for which there exist a set $X \subseteq \cX$ of cardinality $d$ that is shattered by $\cH$. If $\cH$ shatters sets of cardinality arbitrarily large, we say that $\VC{\cH} = \infty$.
\end{definition}

\begin{definition}
  Let $\cX$ be a set. A \ul{learning algorithm for functions $\cX \rightarrow \{0,1\}$} is an algorithm $\hh$ that takes a sample $S \in \left(\cX \times \{0,1\}\right)^*$ and outputs a function $\hh_S: ~ \cX \rightarrow \{0,1\}$. The mapping $S \mapsto \hh_S$ may be randomized.  
\end{definition}

\begin{definition}
  Let $\cX$ be a set, let $\cD \in \distribution{\cX \times \{0,1\}}$, and let $h : ~ \cX \rightarrow \{0,1\}$ be a function. The \ul{$0$-$1$ loss of $\cH$ with respect to $\cD$} is $\loss{\cD}{\cH} = \PPP{(x,y) \sim \cD}{h(x) \neq y}$.
\end{definition}

\begin{definition}
  Let $\cX$ be a set, and let $\cH \subseteq \{0,1\}^\cX$ be a class of functions. The set of \ul{realizable distributions for $\cH$} is 
  \[
    \Realizable{\cH} = \left\{\cD \in \distribution{\cX \times \{0,1\}}: ~ \inf_{h \in \cH}\loss{\cD}{h} = 0 \right\}.
  \]
\end{definition}

\subsection{The VCL Dimension}

\begin{definition}\label{definition:d-vcl-tree}
  Let $\cX$ be a set, let $d \in \bbN$ and $\ell \in \bbN \cup \{0,\infty\}$. A \ul{$d$-VCL tree of depth $\ell$ with respect to $\cX$} is a set 
  \begin{equation}\label{eq:d-vc-tree-def}
    T = \big\{
      \bx_\bu \in \cX^d: ~ \bu \in \{0,1\}^{ds}, ~ s \in \bbN \cup \{0\}, ~ s \leq \ell
    \big\}.
  \end{equation}
  We say that \ul{$T$ is infinite} if it has depth $\ell = \infty$.
\end{definition}

Note that a $d$-VCL tree of depth $0$ is not empty, rather it contains a single node $\bx_\lambda$ where $\lambda$ denotes the empty string.

\begin{definition}
  Let $\cX$ be a set, let $\cH \subseteq \{0,1\}^\cX$ be a hypothesis class, and let $d \in \bbN$.
  For each $s \in \bbN$, let $\bx_s = (x_s^1,\dots,x_s^d) \in \cX^d$ and $\by_s = (y_s^1,\dots,y_s^d) \in \{0,1\}^d$. Let $h \in \cH$. 
  For any $t \in \bbN$, we say that the finite sequence $(\bx,\by)_{\leq t} = \bigl((\bx_s,\by_s)\bigr)_{s = 1}^t$ is \ul{consistent with $h$} if
  \begin{equation}\label{eq:consistent}
    \forall s \in [t] \: \forall i \in [d]: ~ h(x_s^i) = y_s^i.
  \end{equation} 
  We say that the infinite sequence $\bigl((\bx_s,\by_s)\bigr)_{s \in \bbN}$ is \ul{consistent with $\cH$} if for any $t \in \bbN$ there exists $h \in \cH$ such that $(\bx,\by)_{\leq t}$ is \ul{consistent with $h$}.
\end{definition}


\begin{definition}\label{definition:shattering-of-vcl-tree}
  Let $\cX$ be a set, let $\cH \subseteq \{0,1\}^\cX$ be a hypothesis class, and let $T$ be a $d$-VCL tree as in \cref{eq:d-vc-tree-def}. We say that \ul{$\cH$ shatters $T$} if for every $t \in \bbN$, $t \leq \ell$, and every $\by \in \{0,1\}^{dt}$ there exists a hypothesis $h \in \cH$ that is consistent with $\bigl((\bx_{\by_{\leq s-1}}, \by_{s})\bigr)_{s=1}^{t}$ in the sense that 
  \begin{equation}\label{eq:shattered-is-consistent}
    \forall s \in [t] \: \forall j \in [d]: ~ h(x_{\by_{\leq s-1}}^j) = y_{s}^j,
  \end{equation}
  where we use the notation
  \[
    \by_{\leq s} = \Bigl(\left(y_1^1,\dots,y_1^d\right),\dots,\left(y_s^1,\dots,y_s^d\right)\Bigr) \in \{0,1\}^{ds}
  \]
  to denote a prefix of $\by$, and 
  \[
    \bx_{\by_{\leq s}} = \left(x_{\by_{\leq s}}^1,\dots,x_{\by_{\leq s}}^d\right) \in \{0,1\}^{d}
  \]
  to denote the members of $\bx_{\by_{\leq s}}$.
\end{definition}

The $d$-VCL trees used in this paper are a variant of the trees used in \cite{DBLP:conf/stoc/BousquetHMHY21}. To distinguish the two, we call their construction \emph{strong} VCL trees.

\begin{definition}\label{definition:strong-vcl-tree}
  Let $\cX$ be a set, let $d \in \bbN$. An \ul{infinite strong VCL tree with respect to $\cX$} is a set 
  \begin{equation*}\label{eq:strong-vcl-tree-def}
    T = \big\{
      \bx_\bu \in \cX^{s+1}: ~ s \in \bbN\cup\{0\} ~ \land ~ \bu \in \{0,1\}^1 \times \{0,1\}^2 \times \cdots \times \{0,1\}^s
    \big\}.
  \end{equation*}
\end{definition}

\begin{definition}\label{definition:shatter-strong-vcl}
  Let $\cX$ be a set, let $\cH \subseteq \{0,1\}^\cX$ be a hypothesis class, and let $T$ be an infinite strong VCL tree as in \cref{eq:strong-vcl-tree-def}. We say that \ul{$\cH$ shatters $T$} if for every $t \in \bbN$, and every $\by \in \{0,1\}^1 \times \{0,1\}^2 \times \cdots \times \{0,1\}^t$ there exists a hypothesis $h \in \cH$ that is consistent with $\bigl((\bx_{\by_{\leq s-1}}, \by_{s})\bigr)_{s=1}^{t}$ in the sense that 
  \begin{equation}
    \forall s \in [t] \: \forall j \in [s]: ~ h(x_{\by_{\leq s-1}}^j) = y_{s}^j,
  \end{equation}
  where we use the notation
  \[
    \by_{\leq s} = \Bigl(\left(y_1^1\right),\left(y_2^1,y_2^2\right),\left(y_3^1,y_3^2,y^3_3\right),\dots,\left(y_s^1,\dots,y_s^s\right)\Bigr) \in \{0,1\}^{\left(\sum_{k = 1}^s k\right)} 
  \]
  to denote a prefix of $\by$, and 
  \[
    \bx_{\by_{\leq s}} = \left(x_{\by_{\leq s}}^1,\dots,x_{\by_{\leq s}}^{s+1}\right) \in \{0,1\}^{s+1}
  \]
  to denote the members of $\bx_{\by_{\leq s}}$.
\end{definition}

\begin{definition}
  Let $\cX$ be a set and let $\cH \subseteq \{0,1\}^\cX$. The \ul{Vapnik--Chervonenkis--Littlestone dimension of $\cH$}, denoted $\VCL{\cH}$, is the largest integer $d \geq 0$ such that $\cH$ shatters an infinite $d$-VCL tree.
  If $\cH$ does not shatter any infinite $1$-VCL tree, we say that $\VCL{\cH} = 0$. If $\cH$ shatters infinite $d$-VCL trees for $d$ arbitrarily large, we say that $\VCL{\cH} = \infty$.
\end{definition}

\subsection{Learning Rates}

\cite{DBLP:conf/stoc/BousquetHMHY21} used the following definition of distribution-dependent learning rates. 

\begin{definition}[{\citealt[Definition 1.4.]{DBLP:conf/stoc/BousquetHMHY21}}]\label{definition:learning-rates-classic}
  Let $\cH$ be a concept class, and let $R: \mathbb{N} \rightarrow[0,1]$ with $R(n) \rightarrow 0$ be a rate function.
  \begin{itemize}
    \item{
      \ul{$\cH$ is learnable at rate $R$} if there exists a learning algorithm $\hh$ such that for every $\cD \in \Realizable{\cH}$, there exist $C, c \geq 0$ such that $\EEE{S \sim \cD^n}{\loss{\cD}{\hh_S}} \allowbreak  \leq C \cdot R(c \cdot n)$ for all $n \in \bbN$.
    }
    \item{
      \ul{$\cH$ is learnable with rate no faster than $R$} if for every learning algorithm $\hh$, there exists a $\cD \in \Realizable{\cH}$ and $C, c > 0$ for which $\EEE{S \sim \cD^n}{\loss{\cD}{\hh_S}} \geq C \cdot R(c \cdot n)$ for infinitely many $n \in \bbN$.
    }
    \item{
      \ul{$\cH$ is learnable with optimal rate $R$} if $\cH$ is learnable at rate $R$ and $\cH$ is not learnable faster than $R$.
    }
    \item{
      \ul{$\cH$ requires arbitrarily slow rates} if, for every $R(n) \rightarrow 0, \cH$ is learnable at rate no faster than~$R$.
    }
  \end{itemize}
\end{definition}

In this paper we refine the notion of learning rates, introducing the following more nuanced expressions for linear rates, as follows. Note that our definitions are strictly special cases in the sense that if a class is learnable at rate (learnable at rate no faster than) $\sfrac{d}{n}$ according to our definition, then it is learnable at rate (learnable at rate no faster than) $\sfrac{d}{n}$ according to \cref{definition:learning-rates-classic} as well.

\begin{definition}\label{definition:learning-rates-new}
  Let $\cX$ be a set, let $\cH \subseteq \{0,1\}^\cX$ be a hypothesis class, and let $d \geq 0$ and $\gamma \geq 1$. We say that:
  
  \begin{itemize}
    \item{
      \ul{$\cH$ is learnable with fine-grained rate $\sfrac{d}{n}$} if there exists a learning algorithm $\hh$ such that for any distribution $\cD \in \Realizable{\cH}$ there exist real numbers $C,c \geq 0$ such that for all $n \in \bbN$:
      \[
        \EEE{S \sim \cD^n}{\loss{\cD}{\hh_S}} \leq \frac{d}{n} + C\cdot\exp(-cn).
      \]
    }
    \item{
      \ul{$\cH$ is learnable with fine-grained rate no faster than $\sfrac{d}{n}$} if for any learning algorithm $\hh$ there exists a distribution $\cD \in \Realizable{\cH}$ such that the inequality
      \[
        \EEE{S \sim \cD^n}{\loss{\cD}{\hh_S}} \geq \frac{d}{n}
      \]
      holds for infinitely many $n \in \bbN$.
    }
    \item{
      \ul{$\cH$ is learnable with optimal fine-grained rate $\sfrac{d}{n}$ with gap factor $\gamma$} if $\cH$ is learnable with rate no faster than $\sfrac{d}{n}$, and is learnable with rate $\sfrac{d'}{n}$, where $d' \leq \gamma d$.
    }
  \end{itemize}
\end{definition}

To distinguish the two notions of rate, we will refer to the rates of \cref{definition:learning-rates-classic} as \emph{coarse rates}.

\begin{remark}
  Ideally, we would like to obtain a gap factor $\gamma$ that is as close as possible to $1$, so that $d=d'$ (see \cref{definition:learning-rates-new}). The extent to which this is possible is a topic for further research. Throughout this paper we use $\gamma = 800$.
\end{remark}

\begin{definition}\label{definition:strongly-distribution-dependent-linear}
  Let $\cX$ be a set, let $\cH \subseteq \{0,1\}^\cX$ be a hypothesis class. We say that \ul{$\cH$ is learnable with a strongly distribution-dependent linear rate} if for any (possibly randomized) learning algorithm $\hh$ and any $c \geq 0$ there exists $\cD \in \Realizable{\cH}$ such that the inequality
  \begin{equation}
    \frac{ c}{n} \leq \EEE{S \sim \cD^n}{\loss{\cD}{\hh_S}}
  \end{equation}
  holds for infinitely many $n \in \bbN$. 
\end{definition}

\begin{remark}
  There are various technical issues related to measure theory that arise in the distribution-dependent learning setting and are germane to our results. We use the same assumptions as \cite{DBLP:conf/stoc/BousquetHMHY21}, and refer the interested reader to their work for an in-depth discussion (e.g., Section 3.3 and Appendices B and C).
\end{remark}

\subsection{Gale--Stewart Games}

We will use some basic concepts and results concerning infinite games. We refer the reader to Appendix A.1 of \cite{DBLP:conf/stoc/BousquetHMHY21} for additional references and discussion. Briefly, we consider infinite full information two-player games, in which there exists a set $\Omega$ and a subset $W \subseteq \Omega^\bbN$, and at each time $t=1,2,3,\dots,$ Player $1$ selects an item $x_t \in \Omega$, and then Player $2$ selects an item $y_t \in \Omega$. Player $1$ wins if and only if the resulting infinite sequence $\bz = (x_1,y_1,x_2,y_2,\dots)$ satisfies $\bz \in W$; otherwise, Player $2$ wins.

We say that Player $i$ has a \emph{winning strategy} if there exists a function $f: ~ \Omega^* \rightarrow \Omega$ such that if in every time $t \in \bbN$, Player $i$ selects item $f(\bz')$ where $\bz'$ is the finite sequence of all items selected so far (by both players), then Player $i$ wins the game (regardless of the selections made by the other player).

A game is called \emph{determined} if precisely one of the players has a winning strategy. An infinite game is called \emph{Gale--Stewart} (or \emph{finitely-decidable}) if for every $\bw \in W$ there exists $t \in \bbN$ such that for any infinite suffix $\bs \in \Omega^\bbN$, $\bw_{\leq t} \circ \bs \in W$, where `$\circ$' denotes concatenation. Namely, every member of $W$ has a finite prefix that certifies its membership in $W$. We will use the following result.

\begin{theorem}[{\citealt*[][]{GS53}}]\label{theorem:GS-determined}
  Every Gale--Stewart game is determined.
\end{theorem}

\section{Technical Overview}

Our first result is the characterization of fine-grained learning rates via the VCL dimension. 

\begin{theorem}\label{theorem:fine-grained}
  There exist constants $\alpha,\beta > 0$ as follows. Let $\cX$ be a set, let $\cH \subseteq \{0,1\}^\cX$ be a hypothesis class, and let $d = \VCL{\cH}$. 
  \begin{enumerate}
    \item{
      If $d < \infty$ then $\cH$ is learnable with optimal fine-grained rate $\sfrac{d}{n}$ with gap factor $\gamma = \sfrac{\beta}{\alpha}$; furthermore, the marginal of the hard distribution on $\cX$ depends only on $\cH$. Namely, there exists $\cD_\cX \in \distribution{\cX}$ such that for any (possibly randomized) learning algorithm $\hh$ there exists $\cD \in \Realizable{\cH}$ such that the marginal distribution of $\cD$ on $\cX$ is $\cD_\cX$, and the inequality
      \[
        \alpha \cdot\frac{ d}{n} \leq \EEE{S \sim \cD^n}{\loss{\cD}{\hh_S}}
      \]
      holds for infinitely many $n \in \bbN$; and there exists a learning algorithm $h^*$ such that for any $\cD \in \Realizable{\cH}$ there exist parameters $C,c > 0$ such that 
      \[
        \forall n \in \bbN: ~ \EEE{S \sim \cD^n}{\loss{\cD}{h^*_S}} 
        \leq \beta\cdot \frac{d}{n} + C \cdot e^{-c \cdot n}.
      \]
    }
    \item{
      Otherwise, if $\cH$ does not shatter an infinite strong VCL tree, then $\cH$ is learnable with a strongly distribution-dependent linear rate. Namely, for any (possibly randomized) learning algorithm $\hh$ and any $c > 0$ there exists $\cD \in \Realizable{\cH}$ such that the inequality
      \[
        \frac{c}{n} \leq \EEE{S \sim \cD^n}{\loss{\cD}{\hh_S}}
      \]
      holds for infinitely many $n \in \bbN$; and there exists a learning algorithm $h^*$ such that for any $\cD \in \Realizable{\cH}$ there exists $c > 0$ such that
      \[
      \forall n \in \bbN: ~ \EEE{S \sim \cD^n}{\loss{\cD}{h^*_S}} \leq \frac{c}{n}.
      \]
    }
    \item{
      Otherwise, $\cH$ requires arbitrarily slow rates.
    }
  \end{enumerate}
\end{theorem}

\begin{remark}
	Our proofs use $\alpha = \sfrac{1}{100}$, $\beta = 8$, and $\gamma = 800$.
\end{remark}

All proofs appear in \cref{section:proofs,section:half-spaces}. Additionally, we provide a brief overview of the main ideas in each proof.

\begin{figure}[H]
	\centering
	\begin{subfigure}[b]{1\textwidth}
		\centering
		\includegraphics[width=0.5\textwidth]{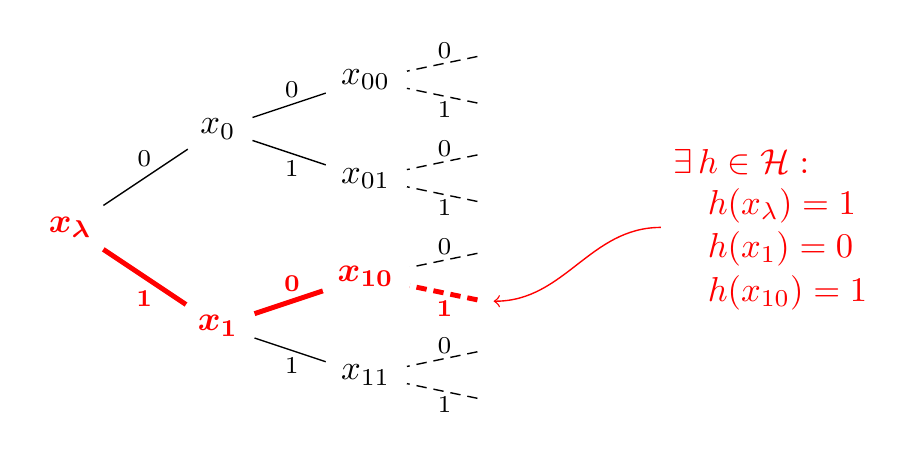}
    
		\caption{\footnotesize
			The first few layers of an infinite $1$-VCL tree (also called an \emph{infinite Littlestone tree}). \hfill \strut
		}
	\end{subfigure}
	\hfill
	\begin{subfigure}[b]{1\textwidth}
		\centering
		\includegraphics[width=0.7\textwidth]{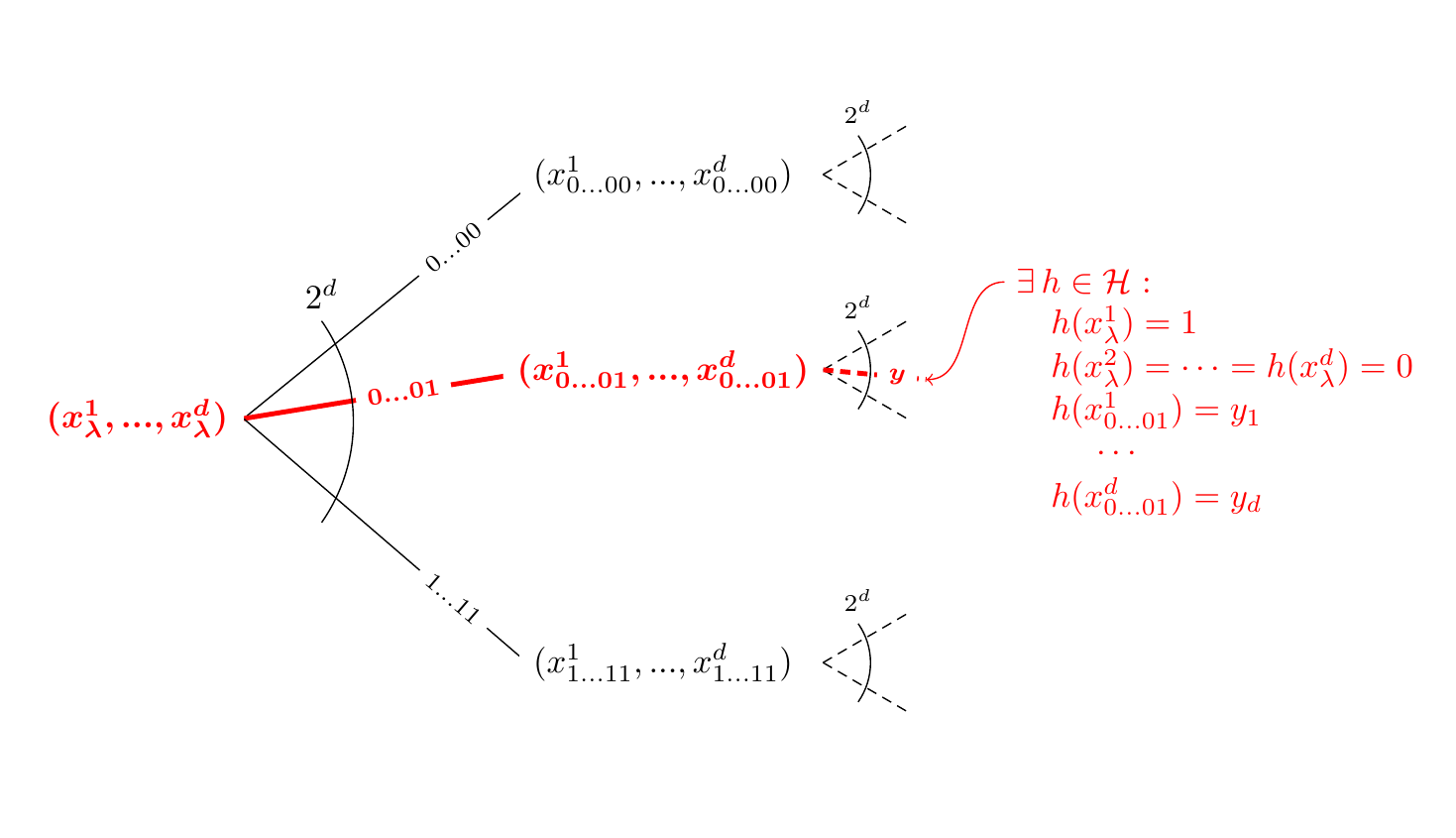}
		\caption{\footnotesize
			The first few layers of an infinite $d$-VCL tree. 
			Every arc represents $2^d$ children, one for each possible labeling of the $d$ points in the preceding node.
			Introduced in this paper, $d$-VCL trees form the basis for our novel characterization. \label{fig:d-vcl}
		}
	\end{subfigure}
	\hfill
	\begin{subfigure}[b]{1\textwidth}
		\centering
		\includegraphics[width=0.77\textwidth]{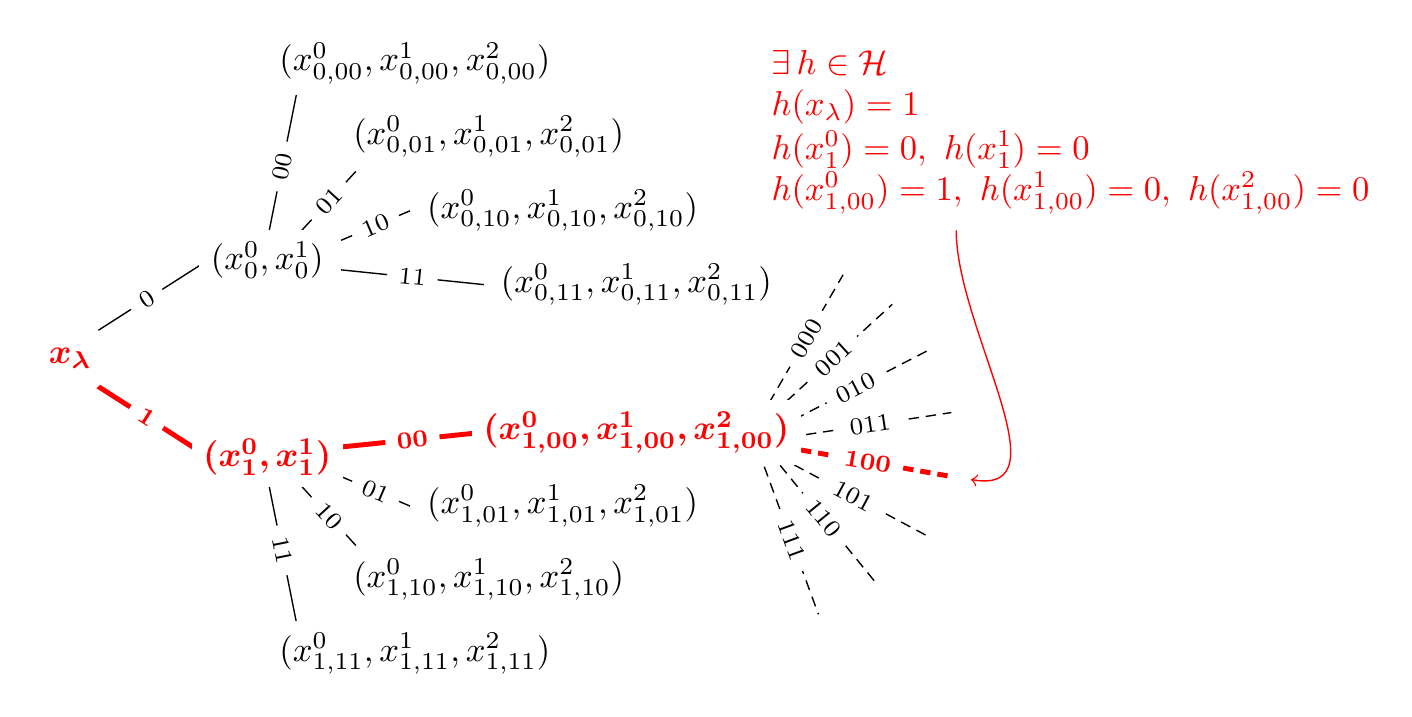}
		\caption{\footnotesize
      The first few layers of an infinite strong VCL tree. 
			(In \cite{DBLP:conf/stoc/BousquetHMHY21} this structure was called an \emph{infinite VCL tree}. We add the modifier \emph{strong} to distinguish it from $d$-VCL trees.)
			\label{fig:strong-vcl}}
	\end{subfigure}
	\caption{\footnotesize
    VCL trees. Every finite branch is consistent with a concept $h\in\mathcal{H}$. 
		This is illustrated here for one branch in each tree, shown in red.\\
		\vspace*{2cm} \hfill {\footnotesize Source: \cite{DBLP:conf/stoc/BousquetHMHY21}, adapted with permission.}}
\end{figure}

The proof of \cref{theorem:fine-grained} is similar to the proof of \cref{theorem:trichotomy} from \cite{DBLP:conf/stoc/BousquetHMHY21}. One of the main differences is that we use $d$-VCL trees for the characterization. Our $d$-VCL trees (introduced in this paper, see \cref{definition:d-vcl-tree,fig:d-vcl}), are an intermediary refinement that lies between the $1$-VCL trees and the strong VCL trees that were used in their proof. Identifying that this particular combinatorial structure characterizes the fine-grained rate is a non-trivial contribution of this paper.

Proving the lower bound of \cref{theorem:fine-grained} requires some technical improvements upon the technique of \cite{DBLP:conf/stoc/BousquetHMHY21}. For each learning algorithm, they constructed a hard distribution that is concentrated on an infinite `target' branch chosen at random in the $d$-VCL tree, and argued that if a test point is deeper in the tree than all points in the training set, then the leaner will make an incorrect prediction on that test point with probability $1/2$. That approach is not suitable for constructing a single marginal distribution $\cD_\cX \in \distribution{\cX}$ that is hard for all learning algorithms (because for every target branch there exists an algorithm that returns a hypothesis with low loss on that branch). Instead, we choose a marginal $\cD_\cX$ that is distributed roughly evenly over all branches in the tree, and construct a distribution $\cD \in \Realizable{\cH}$ that has marginal $\cD_\cX$, and has labels corresponding to an infinite target branch in the tree.

In a general $d$-VCL tree, our approach would be problematic, because even if a test point is deeper in the tree than all the points in the training set, the labels for points in the training set \emph{that do not belong to the target branch} could provide information about the target branch.\footnote{Consider the case where for some $x \in \cX$ there exist infinite branches $\by^{(0)}$ and $\by^{(1)}$ in the tree, both of which do not contain $x$, such that for all $b \in \{0,1\}$, it holds that all $h \in \cH$ that are consistent with $\by^{(b)}$ satisfy $h(x) = b$. Then knowing the label for $x$ allows the learner to eliminate one of the branches.} We overcome this problem as described in the following proof idea.

\begin{proof}[Proof idea for \cref{theorem:fine-grained}]
  For the upper bound, we show that the nonexistence of $d$-VCL trees is equivalent to the existence of a winning strategy for the learner in an infinite two-player game called the `forbidden pattern game'. The equivalence is established via an intermediary online-learning game that is easier to analyze because it is a Gale-Stewart game (whereas the forbidden pattern game is not). A winning strategy for the learner in the forbidden pattern game can be converted into a learning algorithm for the distribution-dependent learning setting, by way of the one-inclusion algorithm of \cite*{haussler:94}. The resulting algorithm has the desired rate of at most $\sfrac{d}{n}$.

  For the lower bound, we use an elementary lemma from Ramsey theory to show that if $\cH$ shatters an infinite $d$-VCL tree, then it also shattered an infinite $d$-VCL tree that satisfies an additional \emph{indifference} property that we define. This property implies that for any infinite branch in the $d$-VCL tree, labels for points that do not belong to the branch provide no information about the labels for points that appear lower down in the tree along the branch. Therefore, when the target branch is chosen randomly, we can argue that if the test point appears on the target branch and is lower than all the training samples, then the learner will make an incorrect prediction with probability $1/2$. The lower bound also involves a specific choice of parameters for the hard marginal distribution that enables a delicate application of Fatou's lemma.
\end{proof}

As a corollary of our characterization, we recover the lower bound of \cite{antos:98} for half-spaces in $\bbR^d$, up to a constants factor.

\begin{theorem}[{\citealt*[][Corollary 1]{antos:98}}]\label{theorem:half-spaces}
  There exists a constant $\alpha > 0$ as follows. Let $d \in \bbN$ and $\cX = \bbR^d$. Let $\cH_d \subseteq \{0,1\}^\cX$ be the set of closed half-spaces in $\bbR^d$. For any learning algorithm $\hh$ there exists a distribution $\cD \in \Realizable{\cH_d}$ such that the inequality
  \[
    \EEE{S \sim \cD^n}{\loss{\cD}{\hh_S}} \geq \alpha\cdot\frac{d}{n}
  \]
  holds for infinitely many values $n \in \bbN$.
\end{theorem}

\begin{proof}[Proof idea]
  It suffices to show that $\VCL{\cH_d} = d$. We consider the dual class for $\cH_d$, and show via a neat `fractal' argument that one can construct an infinite $d$-VCL tree for $\cH_d$. 
\end{proof}

We believe that the `fractal' argument from this proof could be used to construct $d$-VCL trees for other classes as well. 

Finally, we present another, mostly equivalent viewpoint on our results. Stated in a language that is closer to standard PAC learning, it enables a comparison between PAC bounds and distribution-dependent bounds.

\begin{definition}
  Let $\cX$ be a set and let $\cH \subseteq \{0,1\}^\cX$ be a hypothesis class. Let $m: ~ (0,1)^2 \rightarrow \bbN$ and $k: ~ \Realizable{\cH} \rightarrow \bbN$ be functions. We say that \ul{$\cH$ is eventually PAC learnable with sample complexity $m$ and kick-in time $k$} if there exist an algorithm $\hh$ such that for any distribution $\cD \in \Realizable{\cH}$ and for any $\varepsilon,\delta \in (0,1)$ the inequality
  \[
    \PPP{S \sim \cD^n}{\loss{\cD}{\hh_S} \leq \varepsilon} \geq 1-\delta
  \]
  holds for all $n \geq \max\{m(\varepsilon,\delta), k(\cD)\}$.
\end{definition}

\begin{theorem}\label{theorem:pac-style}
  There exist constants $\alpha,\beta > 0$ as follows. Let $\cX$ be a set, let $\cH \subseteq \{0,1\}^\cX$ be a hypothesis class, and let $d \in \bbN$.
  \begin{enumerate}
    \item{
      If $d = \VCL{\cH} < \infty$ then $\cH$ is eventually PAC learnable with sample complexity $m(\varepsilon,\delta) \leq \alpha d\log(\sfrac{1}{\delta})/\varepsilon$.
    }
    \item{
      If $\cH$ is eventually PAC learnable with sample complexity 
      \[
        m(\varepsilon,\delta) = d\log(\sfrac{1}{\delta})/\varepsilon, 
      \]
      then $\VCL{\cH} \leq \beta d$.
    }
  \end{enumerate} 
\end{theorem}

\begin{proof}[Proof idea]
  This follows from \cref{theorem:fine-grained}, together with a standard amplification argument for converting an algorithm with bounded expected error to a PAC learner.
\end{proof}

\section{Proof of the Fine-Grained Characterization}\label{section:proofs}

\subsection{Upper Bound}

Throughout this section, let $\cX$ be a set, let $\cH \subseteq \{0,1\}^\cX$ be a hypothesis class, and let $d \in \bbN$.

\begin{definition}
  The \ul{online learning game for $\cH$ of size $d$}, denoted $\OnlineGame{\cH}{d}$, is an infinite full information game played between two players, a \emph{learner} and an \emph{adversary}. At each time step $t = 1,2,3,\dots$:
  \begin{enumerate}
    \item{
      The adversary chooses $\bx_t = (x_t^1,\dots,x_t^d) \in \cX^d$.
    }
    \item{
      The learner chooses $\by_t = (y_t^1,\dots,y_t^d) \in \{0,1\}^d$.
    }
  \end{enumerate}
  For each $t \in \bbN$, the \ul{version space} is defined by 
  \[
    \cH_t = \cH_{\bx_1,\by_1,\dots,\bx_t,\by_t} = \left\{h \in \cH: \left(\forall s \in [t] \: \forall i \in [d]: ~ h(x_s^i) = y_s^i \right)\right\}.
  \]
  If there exists a time step $t \in \bbN$ such that $\cH_t = \varnothing$ then the learner wins the game. Otherwise, the adversary wins the game.
\end{definition}


\begin{definition}
  The \ul{forbidden pattern game for $\cH$ of size $d$}, denoted $\ForbiddenGame{d}{\cH}$, is an infinite full information game played between two players, a \emph{learner} and an \emph{adversary}. At each time step $t = 1,2,3,\dots$:
  \begin{enumerate}
    \item{
      The adversary chooses $\bx_t \in \cX^d$.
    }
    \item{
      The learner chooses $\hat{\by}_t \in \{0,1\}^d$.
    }
    \item{
      The adversary chooses $\by_t \in \{0,1\}^d$.
    }
  \end{enumerate}
  The adversary wins the game if the adversary's infinite sequence $\bigl((\bx_t,\by_t)\bigr)_{t \in \bbN}$ is consistent with $\cH$ and $\hat{\by}_t = \by_t$ for infinitely many $t \in \bbN$. Otherwise, the learner wins the game. 
\end{definition}

In other words, the learner wins the forbidden pattern game if $\hat{\by}_t$ is eventually a `forbidden pattern' that is not consistent with $\cH$.

We show that the existence of $d$-VCL trees characterizes the winner in the forbidden pattern game. Note that while the online game is a Gale-Stewart game, the forbidden pattern game is not. This makes the online game a convenient stepping stone towards this characterization, as in the following claim.

\begin{lemma}\label{lemma:upper-bound-equivalence}
  The following conditions are equivalent:
  \begin{enumerate}
    \item{\label{item:no-d-vc-tree}
      There does not exist an infinite $d$-VCL tree with respect to $\cX$ that is shattered by $\cH$.
    }
    \item{\label{item:winning-strategy-for-learner-online}
      There exists a winning strategy for the leaner in the online game $\OnlineGame{\cH}{d}$.
    }
    \item{\label{item:winning-strategy-for-learner-forbidden}
      There exists a winning strategy for the leaner in the forbidden pattern game $\ForbiddenGame{\cH}{d}$.
    }
  \end{enumerate}
\end{lemma}


The proof of \cref{lemma:upper-bound-equivalence} is divided between \cref{claim:d-vc-tree-iff-winning-strategy-for-adversary-online,claim:item1-iff-item2,claim:item2-implies-item3,claim:item3-implies-item1}.

\begin{claim}\label{claim:d-vc-tree-iff-winning-strategy-for-adversary-online}
  There exists an infinite $d$-VCL tree with respect to $\cX$ that is shattered by $\cH$ if and only if there exists a winning strategy for the adversary in $\OnlineGame{\cH}{d}$.
\end{claim}

\begin{proof}
  Assume that there exists an infinite $d$-VCL tree
  \begin{equation}\label{eq:infinite-d-vc-tree}
    T = \big\{
      \bx_\bu \in \cX^d: ~ \bu \in \left(\{0,1\}^{d}\right)^*
    \big\}
  \end{equation}
  that is shattered by $\cH$. This implies the existence of a winning strategy for the adversary as following. At each time step $t \in \bbN$, the adversary selects $\bx_t = \bx_{\by_{\leq t-1}}$. For any choice $\by_t \in \{0,1\}^d$ made by the learner the version space remains not empty, i.e., $\cH_t \neq \varnothing$. This holds because $\cH$ shatters $T$, and so in particular there exists a hypothesis $h \in \cH$ that is consistent with $\bigl((\bx_{\by_{\leq s-1}}, \by_{s})\bigr)_{s\in[t]}$. Hence, the adversary wins the game when playing according to this strategy.

  Conversely, assume that there exists a winning strategy for the adversary defined by a function $f: \{0,1\}^* \rightarrow \cX^d$ such that at any time step $t \in \bbN$, the adversary chooses $\bx_t = f(\by_{\leq t-1})$, where $\by_{\leq t-1}$ is the sequence of choices the learner has made so far. The function $f$ defines an infinite $d$-VCL tree $T$ as in \cref{eq:infinite-d-vc-tree} given by $\bx_{\bu} = f(\bu)$. Seeing as this is a winning strategy for the adversary, $\cH_t \neq \varnothing$ for all $t \in \bbN$ and all possible choices of $\by_{\leq t}$, and this implies that the tree $T$ is shattered by $\cH$. 
\end{proof}

\begin{claim}\label{claim:item1-iff-item2}
  In the context of \cref{lemma:upper-bound-equivalence}, \cref{item:no-d-vc-tree} $\iff$ \cref{item:winning-strategy-for-learner-online}. 
\end{claim}

\begin{proof}
  \begin{align*}
    \left(\text{\cref{item:no-d-vc-tree}}\right) ~ &\iff ~ \left(\text{$\nexists$ winning strategy for the adversary in $\OnlineGame{\cH}{d}$}\right) \\
    ~ &\iff ~ 
    \left(\text{\cref{item:winning-strategy-for-learner-online}}\right),
  \end{align*}

  where the first equivalence is by \cref{claim:d-vc-tree-iff-winning-strategy-for-adversary-online}, and the second equivalence states that the online learning game is determined, which is true by \cref{theorem:GS-determined} because it is a Gale--Stewart game.
\end{proof}

\begin{claim}\label{claim:item2-implies-item3}
  In the context of \cref{lemma:upper-bound-equivalence}, \cref{item:winning-strategy-for-learner-online} $\Longrightarrow$ \cref{item:winning-strategy-for-learner-forbidden}. 
\end{claim}

\begin{algorithmFloat}[!ht]
  \begin{ShadedAlgorithmBox}
    {\bf Assumption:} $f: \left(\bigcup_{s = 1}^\infty\cX^{ds}\right) \rightarrow \{0,1\}^d$ is a function that defines a winning strategy for the learner in the online game $\OnlineGame{d}{\cH}$.

    \noindent\rule{\textwidth}{0.5pt}

    \vspace*{0.5em}

    \textsc{ForbiddenPatternLearner}:	
    \begin{algorithmic}
      \State $\xi 
      \gets$ empty sequence
      \State $\eta 
      \gets$ empty sequence
      \For $t \gets 1,2,\ldots:$
      \State Receive $\bx_t$ from the adversary
      \State Choose $\hat{\by}_t \gets f(\xi \circ \bx_t)$
      \State Receive $\by_t$ from the adversary
      \If $\hat{\by}_t = \by_t$:
        \State $\xi \gets \xi \circ \bx_t$
        \State $\eta \gets \eta \circ \by_t$
      \EndIf
      \EndFor
    \end{algorithmic}
  \end{ShadedAlgorithmBox}
  \caption{A reduction from a winning strategy for the forbidden pattern game to a winning strategy for the online game.}
  \label{algorithm:reduction-forbidden-to-online}
\end{algorithmFloat}

\begin{proof}[Proof idea]
  Use \cref{algorithm:reduction-forbidden-to-online}. A winning strategy for the learner in the online game empties the version space. So eventually, for any $\bx_t$ chosen by the adversary, the learner can choose a $\by_t$ such that $(\bx,\by)_{\leq t}$ is not consistent with $\cH$.
\end{proof}

\begin{proof}
  Let $f: \left(\bigcup_{s = 1}^\infty\cX^{ds}\right) \rightarrow \{0,1\}^d$ be a function that defines a winning strategy for the learner in the online game $\OnlineGame{d}{\cH}$. Namely, in the online game, if in each time step $t \in \bbN$ the adversary chooses $\bx_t$ and the learner chooses $\by_t = f(\bx_1,\dots,\bx_t)$, then after a finite number of steps $\cH_t = \varnothing$.

  Given such a function $f$, \cref{algorithm:reduction-forbidden-to-online} defines a winning strategy for the learner in the forbidden pattern game. To see this, assume for contradiction that the strategy of \cref{algorithm:reduction-forbidden-to-online} is not a winning strategy for the learner, namely, assume that there exists a sequence $\bigl((\bx_t,\by_t)\bigr)_{t \in \bbN}$ that is consistent with $\cH$ and also $\hat{\by}_t = \by_t$ for infinitely many $t$ when $\hat{\by}$ is chosen by the learner according to \cref{algorithm:reduction-forbidden-to-online}. This implies that the sequences $\xi$ and $\eta$ defined by the algorithm are infinite. 
  
  We show that if the adversary in the online game plays this infinite sequence $\xi$ and the learner plays according to the strategy $f$, then the adversary wins the game, in contradiction to the assumption that $f$ defines a winning strategy for the learner in the online game. 
  
  Let $\xi = \xi_1,\xi_2,\dots$ and $\eta = \eta_1,\eta_2,\dots$ where $\xi_t \in \cX^d$ and $\eta_t \in \{0,1\}^d$ for all $t \in \bbN$. By construction, $\bigl((\xi_t,\eta_t)\bigr)_{t \in \bbN}$ is consistent with $\cH$ because it is a subsequence of $\bigl((\bx_t,\by_t)\bigr)_{t \in \bbN}$. In particular, for any finite prefix $(\xi,\eta)_{\leq t}$ there exists a hypothesis $h \in \cH$ that is consistent with $(\xi,\eta)_{\leq t}$. This implies that for any $t \in \bbN$, the version space $\cH_t = \cH_{\xi_1,\eta_1,\dots,\xi_t,\eta_t}$ is not empty. However, the sequence $\bigl((\xi_t,\eta_t)\bigr)_{t \in \bbN}$ is constructed by playing according to the strategy $f$, namely $\eta_t = f(\xi_1,\dots,\xi_t)$ for all $t \in \bbN$. We conclude that when the adversary in the online game plays $\xi$ and the learner plays according to $f$, then $\cH_t \neq \varnothing$ for all $t \in \bbN$, yielding the desired contradiction to the choice of $f$.
\end{proof}

\begin{claim}\label{claim:item3-implies-item1}
  In the context of \cref{lemma:upper-bound-equivalence}, \cref{item:winning-strategy-for-learner-forbidden} $\Longrightarrow$ \cref{item:no-d-vc-tree}. 
\end{claim}

\begin{proof}
  We show the contrapositive, namely, if there exists an infinite $d$-VCL tree shattered by $\cH$ then there does not exist a winning strategy for the leaner in the forbidden pattern game $\ForbiddenGame{\cH}{d}$ (this is similar to one of the directions in \cref{claim:d-vc-tree-iff-winning-strategy-for-adversary-online}). Indeed, let $T$ be an infinite shattered tree as in \cref{eq:infinite-d-vc-tree}. Then there exists a winning strategy for the adversary: at each time step $t \in \bbN$, the adversary chooses $\bx_t = \bx_{\by_{\leq t-1}}$, and chooses $\by_t \in \{0,1\}^d$ to be any value such that $\by_t \neq \hat{\by}_t$. Because the tree is shattered, for every $t \in \bbN$ and every possible $\by_t \in \{0,1\}^d$ there exists $h \in \cH$ that is consistent with $(\bx,\by)_{\leq t}$. Hence, the resulting sequence $\bigl((\bx_t,\by_t)\bigr)_{t \in \bbN}$ is consistent with $\cH$ while also satisfying $\by_t \neq \hat{\by}_t$ for all $t \in \bbN$, and therefore the adversary wins the game.
\end{proof}

\begin{notation}\label{notation:learner-strategy-forbidden}
  Fix a function $f$ as in \cref{algorithm:reduction-forbidden-to-online}, and consider an execution of that algorithm using $f$ in which the adversary plays the sequence $\bigl((\bx_t,\by_t)\bigr)_{t \in \bbN}$. For each $t \in \bbN$ let $\xi^{(t)}$ denote the value of $\xi$ at the beginning of time step $t$. We write
  \[
  \hat{\by}_t: ~\cX^d \rightarrow \{0,1\}^d
  \]
  to denote the function given by
  \[
    \hat{\by}_t(x) = \hat{\by}_{(\bx,\by)_{\leq t-1}}(x) = f(\xi^{(t)} \circ x)
  \]
  that determines the learner's choice at time $t$, such that $\hat{\by}_t = \hat{\by}_t(\bx_t)$ for all $t \in \bbN$. 
\end{notation}

\begin{definition}
  Let $\cD \in \Delta(\cX^d \times \{0,1\}^d)$ be a distribution, and let $g: \cX^d \rightarrow \{0,1\}^d$ be a function. The \ul{forbidden pattern loss of $g$ with respect to $\cD$} is 
  \[
    \forbiddenLoss{\cD}{g} = \PPP{(X,Y) \sim \cD}{g(X) = Y} = 1 - \loss{\cD}{g}.
  \]
\end{definition}

The forbidden pattern loss simply captures the learners objective in the forbidden pattern game, which is to avoid having $\hat{\by}_t = \by_t$.

\begin{claim}\label{claim:positive-loss-forbidden-game-vanishes}
  Assume $\VCL{\cH} < \infty$. Let $\cD \in \Realizable{\cH}$ be a distribution, and let $S = \bigl((X_1,Y_1),(X_2,Y_2),\dots\bigr)$ be an infinite sequence of i.i.d.\ samples from $\cD$. Consider an instance of the forbidden pattern game where the adversary plays the sequence $S$, and the learner plays according to the function $\hat{\by}_t = \hat{\by}_{S_{\leq t-1}}$ as in \cref{notation:learner-strategy-forbidden}. Then the forbidden pattern loss satisfies
  \[
    \lim_{t \rightarrow \infty} \PPPunder{S \sim \cD^{\bbN}}{\forbiddenLoss{\cD}{\hat{\by}_t} > 0} = 0.
  \]
\end{claim}

\begin{proof}
  By the proof of \cref{lemma:upper-bound-equivalence} and the assumption that $\VCL{\cH} < \infty$, the strategy $\hat{\by}_t$ is a winning strategy for the learner in the forbidden pattern game.
  
  First, assume that $S$ is consistent with $\cH$. Then there exists a random variable $T \in \bbN$ that depends on $S$, such that 
  \begin{equation}\label{eq:forbidden-learner-wins-for-t-large-enough}
  	\PP{\forall t \geq T: ~ \hat{\by}_t(X_t) \neq Y_t} = 1.
  \end{equation}
  This is true because $\hat{\by}_t$ is a winning strategy for the learner.
  Furthermore, by construction of the strategy $\hat{\by}$, the function $\hat{\by}_t(x)$ only changes if the learner made a mistake, namely
  \begin{equation}\label{eq:forbidden-learner-doesnt-change}
  \PP{\forall t,t' \geq T ~ \forall x \in \cX: ~ \hat{\by}_t(x) = \hat{\by}_{t'}(x)} = 1.
  \end{equation}
  Hence,
  \begin{flalign*}
  	& \lim_{t \rightarrow \infty} \PPPunder{S \sim \cD^{\bbN}}{\forbiddenLoss{\cD}{\hat{\by}_t} = 0} &
  \end{flalign*}
  \vspace*{-1\baselineskip}
  \begin{align*}
  	~~~~ 
  	&= 
  	\lim_{t \rightarrow \infty} \PP{\left(\lim_{K \rightarrow \infty} \frac{1}{K} \sum_{k = 1}^K \1\left(\hat{\by}_{t}(X_{t+k})=Y_{t+k}\right) \right) = 0}
  	\\
  	& \geq 
  	\lim_{t \rightarrow \infty} \PP{\left(\lim_{K \rightarrow \infty} \frac{1}{K} \sum_{k = 1}^K \1\left(\hat{\by}_{t}(X_{t+k})\right) = Y_{t+k}\right) = 0 ~ \land ~ t \geq T}
  	\\
  	&=
  	\lim_{t \rightarrow \infty} \PP{\left(\lim_{K \rightarrow \infty} \frac{1}{K} \sum_{k = 1}^K \1\left(\hat{\by}_{t+k}(X_{t+k})\right) = Y_{t+k}\right) = 0 ~ \land ~ t \geq T}
  	\tag{By \cref{eq:forbidden-learner-doesnt-change}}
  	\\
  	&=
  	\lim_{t \rightarrow \infty} \PP{t \geq T} = 1. \tag{By \cref{eq:forbidden-learner-wins-for-t-large-enough}}
  \end{align*}
  So
  \[
    \lim_{t \rightarrow \infty} \PPPunder{S \sim \cD^{\bbN}}{\forbiddenLoss{\cD}{\hat{\by}_t} > 0} = 1-\lim_{t \rightarrow \infty} \PPPunder{S \sim \cD^{\bbN}}{\forbiddenLoss{\cD}{\hat{\by}_t} = 0} = 0
  \]
  as desired.

  It remains to show that $\PP{\text{$S$ is consistent with $\cH$}} = 1$. This is a consequence of the Borel--Cantelli lemma. Seeing as $\cD \in \Realizable{\cH}$, there exists a sequence $h_1,h_2,\ldots \in \cH$ such that $\loss{\cD}{h_k} \leq 2^{-k}$ for all $k \in \bbN$. For every $t,k \in \bbN$ let $G_{t,k} = \{\forall i \in [t]: ~ h_k(X_i) = Y_i\}$ be the event in which $S_{\leq t}$ is consistent with $h_k$. Then for every $t \in \bbN$,
  \begin{align*}
    \sum_{k \in \bbN} \PPP{S \sim \cD^\bbN}{\neg G_{t,k}} \leq \sum_{k \in \bbN} t \cdot \loss{\cD}{h_k} \leq t < \infty.
  \end{align*}
  By Borel--Cantelli, this implies that 
  \[
    \forall t \in \bbN: ~ \PPP{S \sim \cD^\bbN}{\exists k \in \bbN: ~ G_{t,k}} = 1.
  \]
  In words, for every $t \in \bbN$, with probability $1$ over the choice of $S$, there exists $k \in \bbN$ such that $h_k$ is consistent with $S_{\leq t}$. Finally,
  \begin{align*}
    \PPP{S \sim \cD^\bbN}{\text{$S$ is consistent with $\cH$}} \geq \PPP{S \sim \cD^\bbN}{\bigcap_{t \in \bbN} \left\{\exists k \in \bbN: ~ G_{t,k}\right\}} = 1,
  \end{align*}
  because a countable intersection of probability $1$ events has probability $1$.
\end{proof}

\begin{definition}
  In the context of \cref{claim:positive-loss-forbidden-game-vanishes}, let 
  \[
  t^* = t^*(\cD) = \inf \left(\left\{t \in \bbN: ~ \PPPunder{S \sim \cD^{\bbN}}{\forbiddenLoss{\cD}{\hat{\by}_t} > 0} \leq \fracUB{1}\right\} \cup \{\infty\}\right).
  \]
  The \ul{set of good sample sizes for $\cD$} is 
  \[
    \Tgood{\cD} = \left\{t \in [t^*]: ~ \PPPunder{S \sim \cD^{\bbN}}{\forbiddenLoss{\cD}{\hat{\by}_t} > 0} \leq \fracUB{2}\right\}.
  \]
\end{definition}

\begin{claim}\label{claim:hat-t-works}
  There exists a function $\hat{t}: ~ (\cX \times \{0,1\})^* \rightarrow \bbN$ as follows. Let $\cD \in \Realizable{\cH}$. There exist parameters $C,c \geq 0$ such that for any $n \in \bbN$,
  \[
    \PPP{S \sim \cD^n}{\hat{t}(S) \in \Tgood{\cD}} \geq 1 -Ce^{-cn}.
  \]
\end{claim}

\begin{algorithmFloat}[!ht]
	\begin{ShadedAlgorithmBox}
		{\bf Assumption:} 
		\begin{itemize}
			\item{
				$S = \bigl((X_1,Y_1),\dots,(X_n,Y_n)\bigr) \sim \cD^{n}$ is a labeled training set.
			}
			\item{
				$\hat{\by}_t = \hat{\by}_{S_{\leq t-1}}$ is as in \cref{notation:learner-strategy-forbidden}.
			}
			\item{
				$m = \left\lfloor n/2 \right\rfloor$.
			}
		\end{itemize}
		
		\noindent\rule{\textwidth}{0.5pt}
		
		\vspace*{0.5em}
		
		\textsc{SampleSizeEstimator}:	
		\begin{algorithmic}
			\State $S^{\mathsf{train}},S^{\mathsf{test}} \gets $ independent disjoint subsets of $S$ of size $m$
			\For $t \in [m]$:
				\State $k \gets \left\lfloor m/t \right\rfloor$
				\State $S^{\mathsf{train}}_1,\dots,S^{\mathsf{train}}_k \gets$ independent disjoint subsets of $S^{\mathsf{train}}$ of size $t$
				\For $i \in [k]$:
					\State $\hat{e}_{t,i} \gets \1\left(\exists (X,Y) \in S^{\mathsf{test}}: ~  \hat{\by}_{S^{\mathsf{train}}_i}(X)=Y\right)$
				\EndFor 
				\State $\hat{e}_t \gets \frac{1}{k}\sum_{i \in [k]} \hat{e}_{t,i}$
			\EndFor
			\State $\hat{t} \gets \inf \left(\{t \in [m]: ~ \hat{e}_t \leq \sfrac{3}{16} \} \cup \{\infty\}\right)$
			\State {\bfseries output} $\hat{t}$
		\end{algorithmic}
	\end{ShadedAlgorithmBox}

	\caption{An algorithm for finding $\hat{t}$ such that with high probability, $\hat{t} \in \Tgood{\cD}$.}
	\label{algorithm:sample-size-estimator}
\end{algorithmFloat}

\begin{proof}
  Fix $\cD \in \Realizable{\cH}$. We show that \cref{algorithm:sample-size-estimator} satisfies the requirements of the claim. By \cref{claim:positive-loss-forbidden-game-vanishes}, $t^* = t^*(\cD)$ is finite and $\Tgood{\cD} \neq \varnothing$. 
  
  For each $t \in \bbN$ let $e_t = \PPP{S \sim \cD^{\bbN}}{\forbiddenLoss{\cD}{\hat{\by}_t} > 0}$. Hoeffding's inequality implies that there exist $C_t,c_t \geq 0$ such that $\PP{\left| \hat{e}_t - e_t \right| > \sfrac{1}{16}} \leq C_t\cdot e^{-c_t \cdot n}$. By a union bound, 
  \begin{align}\label{eq:union-bound-for-sample-size-estimator}
  	\PPPunder{S \sim \cD^{\bbN}}{\exists t \in [t^*]: ~ \left| \hat{e}_t - e_t \right| > \sfrac{1}{16}} \leq \sum_{t \in [t^*]} C_t\cdot e^{-c_t \cdot n} \leq C' \cdot e^{-c' \cdot n},
  \end{align}
  for some suitable $C',c' \geq 0$. 
  
  Assume that $m \geq t^*$ and $\forall t \in [t^*]: ~ \left| \hat{e}_t - e_t \right| \leq \sfrac{1}{16}$.
  Then in particular, $\hat{e}_{t^*} \leq e_{t^*} + \sfrac{1}{16} \leq \sfrac{1}{8} + \sfrac{1}{16} = \sfrac{3}{16}$, and therefore the output $\hat{t}$ selected by \cref{algorithm:sample-size-estimator} satisfies $\hat{t} \leq t^*$.
  Additionally, the selected output satisfies $e_{\hat{t}} \leq \hat{e}_{\hat{t}} + \sfrac{1}{16} \leq \sfrac{3}{16} + \sfrac{1}{16} = \sfrac{1}{4}$.
  
  Combining the last paragraph with \cref{eq:union-bound-for-sample-size-estimator}, we conclude that there exist $C,c \geq 0$ such that with probability at least $1-Ce^{-c n}$, $\hat{t}$ satisfies $\hat{t} \leq t^*$ and $e_{t^*} \leq e_{\hat{t}} \leq \sfrac{1}{4}$, so in particular $\hat{t} \in \Tgood{\cD}$, as desired.
\end{proof}

\begin{theorem}[{\citealt*[][Theorem 2.3]{haussler:94}}]\label{theorem:one-inclusion}
  Let $\cF \subseteq \{0,1\}^\cX$ be a hypothesis class. There exists a function
  \[
    A: ~ (\cX \times \{0,1\})^* \times \cX \rightarrow \{0,1\}
  \]
  such that for any target function $f \in \cF$, any $n \in \bbN$, and any $(x_1,\dots,x_n) \in \cX^n$, $A$ satisfies  
  \[
    \frac{1}{|\Symmetric{n}|}\sum_{\sigma \in \Symmetric{n}} L_{\sigma,f}(A) \leq \frac{\VC{\cF}}{n},
  \]
  where $\Symmetric{n}$ is the set of all permutation functions $[n] \rightarrow [n]$, and $L_{\sigma,f}(A)$ is the $0$-$1$ loss of $A$ with respect to $f$ and the permutation $\sigma$, namely,
  \[
    L_{\sigma,f}(A) = \1\!\Big(A\big(x_{\sigma(1)}, f(x_{\sigma(1)}), \dots, x_{\sigma(n-1)}, f(x_{\sigma(n-1)}), x_{\sigma(n)})\big) \neq f(x_{\sigma(n)})\Big).
  \]
\end{theorem}

\begin{claim}\label{claim:one-inclusion-learner}
  For any pattern avoidance function $g:\cX^d \rightarrow \{0,1\}^d$ there exists a function $A_g$ given by
  \[
    A_g: ~ (\cX \times \{0,1\})^* \rightarrow \{0,1\}^{\cX}
  \]
  such that for any distribution $\cD \in \distribution{\cX^d \times \{0,1\}^d}$ for which  $\forbiddenLoss{\cD}{g} = 0$ and for any $n \in \bbN$,
  \[
    \EEE{S\sim \cD^n}{\loss{\cD}{A_g(S)}} \leq \frac{d}{n}.
  \]
\end{claim}

\begin{proof}
  This follows from \cref{theorem:one-inclusion}, along with an appropriate definition of a VC class from $g$.

  Let 
  \[
    \cF = \left\{f \in \{0,1\}^\cX: ~ \left(\forall \bx \in \cX^d: ~ \big( f(\bx_1),\dots,f(\bx_d) \big)\neq g(\bx)\right) \right\}
  \]
  be the set of all functions that avoid the pattern $g(\bx)$ for all $\bx \in \cX^d$. Note that $\VC{\cF} \leq d$ because there does not exist a shattered subset of $\cX$ of cardinality $d$. Let $A_g$ be the function $A$ corresponding to $\cF$ whose existence is guaranteed by \cref{theorem:one-inclusion}. Then
  \begin{flalign}
    & 
    \EEE{S\sim \cD^n}{\loss{\cD}{A_g(S)}} 
    & \nonumber\\ &
    \qquad = \EEE{S\sim \cD^n,(X,Y)\sim \cD}{\1\left((A_g(S))(X) \neq Y\right)} 
    & \nonumber\\ &
    \qquad = \EEE{\big((X_1,Y_1),\dots,(X_{n+1},Y_{n+1})\big)\sim \cD^{n+1},\sigma \sim \uniform{\Symmetric{n+1}}}{L_{\sigma,f}(A_g)}
    & \label{eq:definition-f-labeling-func}\\ &
    \qquad = \EEE{\big((X_1,Y_1),\dots,(X_{n+1},Y_{n+1})\big)\sim \cD^{n+1}}{\frac{1}{|\Symmetric{n+1}|}\sum_{\sigma \in \Symmetric{n+1}} L_{\sigma,f}(A_g)}
    & \nonumber\\ &
    \qquad \leq \frac{\VC{\cF}}{n+1} \leq \frac{d}{n+1}.
    \nonumber
    \tag{By \cref{theorem:one-inclusion}}
    &
  \end{flalign}
  In \cref{eq:definition-f-labeling-func}, $f$ a function in $\cF$ that is consistent with $(X_1,Y_1),\dots,(X_{n+1},Y_{n+1})$. $f$ is chosen deterministically as a function of $(X_1,Y_1),\dots,(X_{n+1},Y_{n+1})$. Such an $f$ exists because $\forbiddenLoss{\cD}{g} = 0$, and $\cF$ contains all functions that avoid $g$. In \cref{eq:definition-f-labeling-func}, we have used the fact that 
  \[
    \bigl(S,(X,Y)\bigr) \stackrel{d}{=} \bigl((X_{\sigma(1)},Y_{\sigma(1)}),\dots,(X_{{\sigma(n+1)}},Y_{{\sigma(n+1)}})\bigr). \qedhere
  \]
\end{proof}

\begin{algorithmFloat}[!ht]
  \begin{ShadedAlgorithmBox}
    {\bf Assumptions:}
    \begin{itemize}[leftmargin=2em]
      \item $n \in \bbN$, $m = \left\lfloor \frac{n}{2}\right \rfloor$.
      \item{$\cD \in \Realizable{\cH}$.}
      \item{$S = \bigl((X_1,Y_1),\dots,(X_n,Y_n)\bigr) \sim \cD^{n}$ is a labeled training set.}
      \item{$x$ is the input that needs to be labeled.}
      \item{$\hat{t}$ is a function as in \cref{claim:hat-t-works}. 
      \item For any sequence $\bz \in \left(\cX^d \times \{0,1\}^d\right)^*$, $\hat{\by}_{\bz}: \cX^d \rightarrow \{0,1\}^d$ is a pattern avoidance function as in \cref{notation:learner-strategy-forbidden}.}
      \item{$A_g$ is a learning algorithm that uses pattern avoidance function $g$, as in \cref{claim:one-inclusion-learner}.}
    \end{itemize}

    \noindent\rule{\textwidth}{0.5pt}

    \vspace*{0.5em}

    \textsc{OptimalRateLearner}$(S)$:	
    \begin{algorithmic}
      \State $\hat{t} \gets \hat{t}(S)$
      \State $k \gets \left\lfloor\sfrac{m}{\hat{t}}\right\rfloor$
      \State $S_{g},S_{a} \gets$ partition of $S$ into two disjoint sets of size at least $m$
      \State $S_1,\dots,S_k \gets$ partition of $S_{g}$ into $k$ disjoint sets of size at least $\hat{t}$
      \For $i \in [k]$:
        \State $g_i \gets \hat{\by}_{S_i}$
        \State $a_i \gets A_{g_i}(S_a)$
      \EndFor
      \State $\hh \gets \bigg(x \mapsto \Majority{a_1(x),\dots,a_k(x)}\bigg)$
      \Comment Defining a function $\hh: \cX \rightarrow \{0,1\}$
      \State {\bfseries output} $\hh$
    \end{algorithmic}
  \end{ShadedAlgorithmBox}
  \caption{An algorithm that achieves the optimal learning rate for any class with finite VCL dimension.}
  \label{algorithm:learning-with-optimal-rate}
\end{algorithmFloat}



\begin{lemma}\label{lemma:forbidden-implies-linear-rate}
  If there exists a winning strategy for the leaner in the forbidden pattern game $\ForbiddenGame{\cH}{d}$, then $\cH$ is learnable with rate $\sfrac{d}{n}$.
\end{lemma}

\begin{proof}[Proof of \cref{lemma:forbidden-implies-linear-rate}]
  Let $\cD \in \Realizable{\cH}$. We need to show that there exist $C,c \geq 0$ as follows. For any $n \in \bbN$, let $\hh_S = \allowbreak \textsc{OptimalRateLearner}(S)$ with $S \sim \cD^n$. Then \[
    \EEE{S \sim \cD^n}{\loss{\cD}{\hh_S}} \leq \sfrac{d}{n} + Ce^{-cn}.
  \]

  This is established via the following analysis of \cref{algorithm:learning-with-optimal-rate}. By \cref{claim:hat-t-works}, there exist $C_0,c_0 \geq 0$ such that 
  \begin{equation}\label{eq:t-hat-good-whp}
    \PPP{S \sim \cD^n}{\hat{t} \in \Tgood{\cD}} \geq 1 -C_0\cdot e^{-c_0\cdot n}.
  \end{equation}
  From the definition of $\Tgood{\cD}$, if $\hat{t} \in \Tgood{\cD}$ then for every $i \in [k]$, 
  \[
    \PPPunder{S \sim \cD^n}{\forbiddenLoss{\cD}{g_i} > 0} \leq \sfracUB{2}.
  \]
  By Hoeffding's inequality, there exist $C_1,c_1 \geq 0$ such that
  \begin{equation}\label{eq:most-gi-are-good-conditional}
    \PPPunder{S \sim \cD^n}{ \frac{\left|\left\{i \in [k]: ~ \forbiddenLoss{\cD}{g_i} > 0\right\}\right|}{k}  \geq \fracUB{3} ~ \Big\vert ~ \hat{t} \in \Tgood{\cD}} \leq C_1\cdot e^{-c_1 \cdot n},
  \end{equation}
  where we have used the fact that if $\hat{t} \in \Tgood{\cD}$ then $k = \Omega(n)$. Applying the inequality $\PP{E} \leq \PP{E | F} + \PP{\neg F}$ to \cref{eq:t-hat-good-whp,eq:most-gi-are-good-conditional} implies that there exist $C,c \geq 0$ such that 
  \begin{equation}\label{eq:most-gi-are-good}
    \PPPunder{S \sim \cD^n}{ \frac{\left|\left\{i \in [k]: ~ \forbiddenLoss{\cD}{g_i} > 0\right\}\right|}{k}  \geq \fracUB{3}} \leq Ce^{-cn},
  \end{equation}

  From \cref{claim:one-inclusion-learner}, for any $i \in [k]$, if $\forbiddenLoss{\cD}{g_i} = 0$ then
  \begin{equation}\label{eq:good-gi-implies-good-ai}
    \EEE{S\sim \cD^n}{\loss{\cD}{A_{g_i}(S)}} \leq \frac{d}{n}.
  \end{equation}

  Let $B$ be the bad event whose probability is bounded by \cref{eq:most-gi-are-good}. Then
  \begin{align}\label{eq:expected-loss-expression}
    \EEE{S\sim \cD^n}{\loss{\cD}{\hh_S}} &= \EEE{S\sim \cD^n}{\loss{\cD}{\hh_S} \cdot \1(B)} + \EEE{S\sim \cD^n}{\loss{\cD}{\hh_S} \cdot \1(\neg B)} \nonumber\\
    &\leq \PPP{S\sim \cD^n}{B} + \EEE{S\sim \cD^n}{\loss{\cD}{\hh_S} \cdot \1(\neg B)} \nonumber\\
    &\leq
    Ce^{-cn} + 
    \EEE{S\sim \cD^n}{\loss{\cD}{\hh_S} \cdot \1(\neg B)},
  \end{align}
  where the final inequality follows by \cref{eq:most-gi-are-good}.

  The expectation in the previous line can be bounded by 
  \begin{flalign}\label{eq:upper-bound-main}
    & \EEE{S\sim \cD^n}{\loss{\cD}{\hh_S} \cdot \1(\neg B)} 
    & \nonumber\\ &
    \qquad = \EEE{S\sim \cD^n, (X,Y) \sim \cD}{\1\left(\Majority{a_1(X),\dots,a_k(X)} \neq Y\right) \1(\neg B)}
    & \nonumber\\ &
    \qquad \leq 
    \PPP{S\sim \cD^n, (X,Y) \sim \cD}{\Majority{a_1(X),\dots,a_k(X)} \neq Y ~ \land ~ \neg B} 
    & \nonumber\\ &
    \qquad \leq 
    \PP{\left(\frac{|\{i: ~ a_i(X) \neq Y\}|}{k} \geq \frac{1}{2}\right)  \land  \left(\frac{|\{i: ~ \forbiddenLoss{\cD}{g_i} = 0\}|}{k} \geq \fracUB{\denomUB - 3}\right)} 
    & \nonumber\\ &
    \qquad \leq 
    \PP{\left(\frac{|\{i: ~ (a_i(X) \neq Y)\land(\forbiddenLoss{\cD}{g_i} = 0)\}|}{k} \geq \fracUB{\halfDenomUB-3}\right)} 
    & \nonumber\\ &
    \qquad \leq
    \frac{8}{k}\cdot \EE{|\{i: ~ (a_i(X) \neq Y)\land(\forbiddenLoss{\cD}{g_i} = 0)\}|} \tag{Markov's inequality}
    & \nonumber\\ &
    \qquad =
    \frac{8}{k}\cdot \sum_{i \in [k]}\PP{(a_i(X) \neq Y)\land(\forbiddenLoss{\cD}{g_i} = 0)}
    & \nonumber\\ &
    \qquad \leq \frac{8}{k}\cdot \sum_{i \in [k]} \frac{d}{n} = \frac{8d}{n},
  \end{flalign}
  where the last inequality follows from \cref{eq:good-gi-implies-good-ai}.

  Finally, plugging the bound of \cref{eq:upper-bound-main} into \cref{eq:expected-loss-expression} yields
  \[
    \EEE{S\sim \cD^n}{\loss{\cD}{\hh_S}} \leq
    Ce^{-cn} + 
    \frac{8d}{n},
  \]
  as desired.~\qedhere
\end{proof}

\subsection{Lower Bound}

\begin{lemma}\label{lemma:lower-bound}
	For any set $\cX$ and any hypothesis class $\cH \subseteq \{0,1\}^\cX$ satisfying $d = \VCL{\cH}$ with $1 \leq d < \infty$, there exists a distribution $\cD_\cX \in \distribution{\cX}$ such that for any (possibly randomized) learning algorithm $\hh$ there exists $\cD \in \Realizable{\cH}$ such that the marginal distribution of $\cD$ on $\cX$ is $\cD_\cX$, and the inequality
	\begin{equation}\label{eq:lower-bound}
	\EEE{S \sim \cD^n}{\loss{\cD}{\hh_S}} \geq \frac{d}{100 \cdot n}
  \end{equation}
	holds for infinitely many $n \in \bbN$.
\end{lemma}

\subsubsection{Ingredients}

The proof employs a claim about \emph{indifferent} $d$-VCL trees, which is proved using a simple lemma from Ramsey theory.

\begin{notation}
  For any $\bu \in \left(\{0,1\}^d\right)^*$, let $\indexFunc{\bu} \in \bbN$ denote the index of $\bu$ in the lexicographical ordering of $\left(\{0,1\}^{d}\right)^*$.
\end{notation}

\begin{definition}
  Let $d \in \bbN$, let $\cX$ be a set, let $\cH \subseteq \{0,1\}^\cX$ be a hypothesis class, and let 
  \[
    T = \left\{ \bx_\bu \in \cX^d: ~ \bu \in \left(\{0,1\}^d\right)^* \right\}
  \]
  be an infinite $d$-VCL tree that is shattered by $\cH$. Recall that this implies the existence of a collection  
  \[
    \cH_T = \left\{h_\bu \in \cH: ~ \bu \in \left(\{0,1\}^d\right)^* \right\}
  \]
  of consistent functions, namely, for each $\bu \in \left(\{0,1\}^d\right)^*$, $h_\bu$ is consistent with the path from the root to node $\bu$, as in the definition of shattering a VCL tree (\cref{definition:shattering-of-vcl-tree}). 
  
  We say that such a collection $\cH_T$ is \ul{indifferent} if for every $\bv,\bu,\bw \in \left(\{0,1\}^d\right)^*$, if $\indexFunc{\bv} < \indexFunc{\bu}$, and $\bw$ is a descendant of $\bu$ in the tree $T$, then $h_\bu(\bx_\bv^j) = h_{\bw}(\bx_\bv^j)$ for every $j \in [d]$. In words, the functions for all the descendants of a node that appears after $\bv$ agree on $\bv$. 

  We say that $T$ is \ul{indifferent} if it has a set $\cH_T$ of consistent functions that are indifferent.
\end{definition}

Intuitively, if $T$ is indifferent, then the labels for a node $\bv$ provide no information on the labels of a node $\bu$ that appears after $\bv$ in the lexicographical order.

The claim about indifferent $d$-VCL trees is as follows.

\begin{claim}\label{claim:indifferent-wlog}
	Let $d \in \bbN$, let $\cX$ be a set, let $\cH \subseteq \{0,1\}^\cX$ be a hypothesis class, and let $T$ be an infinite $d$-VCL tree that is shattered by $\cH$.
  Then there exists an infinite $d$-VCL tree $T'$ that is shattered by $\cH$ that is indifferent.
\end{claim}

Following is the lemma from Ramsey theory used for proving \cref{claim:indifferent-wlog}, and a generalized notion of a trees and subtrees used in that lemma.

\begin{definition}
	Let $(X, \preceq)$ be a partial order relation. For $a,b \in X$, we say that $b$ is a \underline{child} of $a$ if $a \preceq b$ and there does not exist $c \in X$ such that $a \preceq c \preceq b$. For $k \in \bbN$, we say that $(X, \preceq)$ is an \underline{infinite $k$-ary tree} if every $a \in X$ has precisely $k$ distinct children. We say that a partial order $(X', \preceq')$ is a \underline{subtree} of $(X, \preceq)$ if $X' \subseteq X$, and $\forall a,b \in X': ~ a \preceq' b \iff a \preceq b$.
\end{definition}

\begin{lemma}\label{lemma:ramsey}
	Let $T = (X, \preceq)$ be an infinite $k$-ary tree, and let $g: ~ X \rightarrow \{0,1\}$ be a two-coloring of $T$. Then $T$ has a monochromatic infinite $k$-ary subtree $T' = (X', \preceq')$, namely there exists $T'$ such that $T'$ is a subtree of $T$, $T'$ is an infinite $k$-ary tree, and $|g(X')| = |\{g(a): a \in X'\}| = 1$.
\end{lemma}

\begin{proof}[Proof of Lemma~\ref{lemma:ramsey}]
	If there exists $a \in X$ such that the set $X'$ consisting of $a$ and all its descendants satisfies $g(X') = \{1\}$, then we are done (take $T'$ to be the subtree consisting of $a$ and all its descendants). Otherwise, every $a \in X$ has a descendant $b \in X$ such that $g(b) = 0$. This implies that one can construct an infinite $k$-ary subtree that is $0$-monochromatic using the following recursive procedure. Let $r$ be any member of $X$ such that $g(r) = 0$. Let $T'$ be an empty tree, and add $r$ to $T'$. Subsequently, for each node $n$ added to $T'$ (including $r$), for each child $a$ of $n$, add to $T'$ an arbitrary descendant $b$ of $a$ such that $g(b) = 0$.
\end{proof}

\begin{proof}[Proof of Claim~\ref{claim:indifferent-wlog}]
	First, observe that if $T  = \left\{x_\bu: ~ \bu \in \left(\{0,1\}^d\right)^*\right\}$ is an infinite $d$-VCL tree that is shattered by $\cH$ with a collection $\{h_\bu:\: ~ \bu \in \left(\{0,1\}^d\right)^*\}$ of consistent functions, then for any $x \in \cX$ there exists an infinite $d$-VCL tree that is shattered by $\cH$ that is a subtree of $T$ and has a collection of consistent functions that agree on $x$. Indeed, this follows from \cref{lemma:ramsey} by choosing a two-coloring $g: \left(\{0,1\}^d\right)^* \rightarrow \{0,1\}$ of $T$ given by $g(\bu) = h_\bu(x)$.

	Second, we use the above observation to construct an infinite $d$-VCL tree $T' = \left\{x'_\bu: ~ \bu \in \left(\{0,1\}^d\right)^*\right\}$ that is shattered by $\cH$ and is indifferent. The construction works by starting with $T' := T$ and then repeatedly modifying $T'$, as specified in Algorithm~\ref{algorithm:indifferent-tree}.
	Each modification step replaces a subtree $T'_{\bu}$ of $T'$ with one of its own infinite $d$-VCL subtrees, which is obtained by invoking the above observation on $T'_{\bu}$ and $x = \bx_{\bv}^j$ for some $j \in [d]$ and some $\bv$ that precedes $\bu$. In each step, the set of nodes of $T'$ decreases (is replaced by one of its subsets), and the collection of consistent functions can be decreased in a corresponding manner (be replaced by a subset of itself that corresponds to the new set of nodes).
	\begin{algorithmFloat}[H]
		\begin{ShadedAlgorithmBox}
			\begin{algorithmic}
				\State $T' \gets T$
				\For $\bu \in \left(\{0,1\}^d\right)^*$ in lexicographic order:
          \For $\bv \in \left(\{0,1\}^d\right)^*$ such that $\indexFunc{\bv} < \indexFunc{\bu}$:
            \For $j \in [d]$:
              \State replace $T'_\bu$ with an infinite $2^d$-ary subtree of $T'_\bu$ that has
              \State \qquad a collection of consistent functions that agree on $\bx_\bv^j$
            \EndFor
          \EndFor
				\EndFor
			\end{algorithmic}
		\end{ShadedAlgorithmBox}
		\caption{Construction of an indifferent $d$-VCL tree. ($T'_\bu$ denotes the infinite $2^d$-ary subtree of $T'$ rooted at node $\bu$.)}
		\label{algorithm:indifferent-tree}
	\end{algorithmFloat}

	\cref{algorithm:indifferent-tree} never terminates, but it defines an infinite $d$-VCL tree $T'$. $T'$ is well-defined because for every $\br \in \left(\{0,1\}^d\right)^*$, the value of $\bx'_\br$ never changes after the outer loop advances past $\br$ (i.e., $\indexFunc{\bu} > \indexFunc{\br}$), and so $\bx'_\br$ is eventually fixed. $T'$ is an infinite $2^d$-ary subtree of $T$ (each replacement maintains
  that $T'$ is an infinite $2^d$-ary subtree of $T$, so the resulting tree defined by this process is also an infinite $2^d$-ary subtree of $T$). This implies that it is a $d$-VCL tree that is shattered by $\cH$. $T'$ is indifferent by construction, because for each $\bq,\br,\bs \in \left(\{0,1\}^d\right)^*$ and $k \in [d]$, if $\indexFunc{\bq} < \indexFunc{\br}$, and $\bs$ is a descendant of $\br$, then during the iteration of the innermost loop in which $\bu = \br$, $\bv = \bq$, and $j = k$, the subtree $T'_\br$ was replaced with a subtree that has a collection of consistent functions that agree on $(\bx_\bq')^k$. In particular this implies that $h_\br((\bx_\bq')^k) = h_\bs((\bx_\bq')^k)$. This agreement continues to hold from that point onwards, because the collection of consistent functions for descendants of $h_\br$ can only decrease at each step. 
\end{proof}

When a tree is indifferent, it admits a notion of \emph{branch functions}, as follows.

\begin{notation}
  $\cY = \left(\{0,1\}^{d}\right)^{\bbN}$.
\end{notation}

\begin{definition}
  Let $d \in \bbN$, let $\cX$ be a set, let $\cH \subseteq \{0,1\}^\cX$ be a hypothesis class, and let 
  \[
    T = \left\{ \bx_\bu \in \cX^d: ~ \bu \in \left(\{0,1\}^d\right)^* \right\}
  \]
  be an infinite $d$-VCL tree that is shattered by $\cH$ with a collection  
  \[
    \cH_T = \left\{h_\bu \in \cH: ~ \bu \in \left(\{0,1\}^d\right)^* \right\}
  \]
  of consistent functions that are indifferent. 
  Let 
  \[
    \cX_T = \{\bx_\bu^i: ~ \bu \in \left(\{0,1\}^{d}\right)^* ~ \land ~ i \allowbreak \in [d]\}.
  \]
  For every $\by \in \cY$, the \ul{branch function for $\by$} is the unique function $f_\by: ~ \cX_T \rightarrow \{0,1\}$ such that for each $\bv \in \left(\{0,1\}^d\right)^*$ and $j \in [d]$,
  \[
    f_{\by}(\bx_\bv^j) = h_\bu(\bx_\bv^j)
  \]
  for a node $\bu$ such that $\by_{\leq |\bu|} = \bu$ and $\indexFunc{\bu} > \indexFunc{\bv}$. In words, $f_{\by}(\bx_\bv^j)$ is the value assigned to $\bx_\bv^j$ by the consistent function of any node on the infinite branch $\by$ that appears after $\bv$ in lexicographic order. (Due to the indifference property, $h_\bu(\bx_\bv^j)$ is the same for any such node $\bu$.)
\end{definition}

We note some consequences of the definitions of indifference and branch functions.

\begin{claim}\label{claim:properties-of-branches}
  Let $T$ be an indifferent infinite $d$-VCL tree with a collection of branch functions $\{f_\by\}_{\by \in \cY}$. Then:
  \begin{enumerate}
    \item{\label{item:finitely-realizable}
      Every branch function $f_\by$ is \emph{finitely realizable}, meaning that for any finite set $\{x_1,\dots x_m\} \subseteq \cX_T$, there exists a function $h \in \cH$ such that for all $i \in [m]$, $f_\by(x_i) = h(x_i)$.
    }
    \item{\label{item:unique-elements}
      Each element in $T$ is unique. Namely, for every $\bu,\bv \in \left(\{0,1\}^{d}\right)^*$ and every $i,j \in [d]$, if $\bu \neq \bv$ or $i \neq j$ then $\bx_\bu^i \neq \bx_\bv^j$.
    }
    \item{\label{item:indifference-property-branches}
      Let $\bv,\bu \in \left(\{0,1\}^{d}\right)^*$. If $\indexFunc{\bu} > \indexFunc{\bv}$ then there exists $\bb \in \{0,1\}^d$ such that for any $\by \in \cY$, if $\bu = \by_{\leq |\bu|}$ then $f_\by(\bx_\bv^j) = \bb_j$ for all $j \in [d]$. In words, if $\bv$ precedes $\bu$ in lexicographical order, then all the branch functions for branches that pass through node $\bu$ agree on node $\bv$.
    }
  \end{enumerate}
\end{claim}

We think of \cref{item:indifference-property-branches} as an indifference property for branch functions. Intuitively, it means that knowing the labels for $\bv$ does not provide any information on which of the branch functions for branches that pass through $\bu$ is more likely to be the correct labeling function. The branch functions that pass through $\bu$ are \emph{indifferent} to the labels of $\bv$.

\begin{proof}[Proof of \cref{claim:properties-of-branches}]
  \cref{item:finitely-realizable} is immediate from the definition of $f_\by$. For \cref{item:unique-elements}, clearly if $\bu = \bv$ then $\bx_\bu^i \neq \bx_\bv^j$, since otherwise node $\bu$ could not have $2^d$ children, in contradiction to $T$ being a $d$-VCL tree. Assume for contradiction that $\indexFunc{\bv} < \indexFunc{\bu}$ and $\bx_\bu^i = \bx_\bv^j$. Then all consistent functions for the children of $\bu$ must agree on $\bx_\bv^j$, but that implies that they agree on $\bx_\bu^i$ as well, which is again a contradiction to $\bu$ having $2^d$ children. Finally, \cref{item:indifference-property-branches} is immediate from the definition of $f_\by$ and from the indifference of $T$.
\end{proof}

The proof of the lower bound also employs the reverse Fatou's lemma. 

\begin{lemma}[{Reverse Fatou; e.g., \citealt[][Theorem 10.17]{browder1996}\footnote{See also \citetalias{wiki:FatousLemma}.}}]\label{lemma:reverse-fatou}
Let $(\Omega,\cF,\mu)$ be a measure space. Let $g: \Omega \rightarrow \bbR$ be a non-negative measurable function such that $\int_\Omega g \: d\mu < \infty$. For each $n \in \bbN$ let $f_n: ~ \Omega \rightarrow \bbR$ be a measurable function such that $\forall \omega \in \Omega: ~ f_n(\omega) \leq g(\omega)$. Then
\[
  \int_\Omega \limsup_{n \rightarrow \infty} f_n \: d\mu \geq \limsup_{n \rightarrow \infty} \int_\Omega  f_n \: d\mu.
\]
\end{lemma}

\subsubsection{Proof of Lower Bound}

\begin{proof}[Proof of \cref{lemma:lower-bound}]
	We will define a set of distributions 
  \[
    \left\{P_{\by}\right\}_{\by \in \cY} \subseteq \Realizable{\cH}
  \]
  that depends on $\cH$ such that all the distributions in the set have the same marginal distribution over $\cX$. The proof uses the probabilistic method to show that for every learning algorithm $\hh$ for $\cH$ there exists $\by^* \in \cY$ (that depends on $\hh$) such that $P_{\by^*}$ is a hard distribution for $\hh$, namely, that \cref{eq:lower-bound} holds for $\cD=P_{\by^*}$ for infinitely many values of $n$. 
	
	The set $\left\{P_{\by}\right\}_{\by \in \cY}$ is defined as follows. By \cref{claim:indifferent-wlog} and the assumption that $\VCL{\cH} = d$, there exist an indifferent infinite $d$-VCL tree 
	\[
		T = \left\{ \bx_\bu \in \cX^d: ~ \bu \in \left(\{0,1\}^{d}\right)^* \right\}
	\]
  with a corresponding collection of branch functions 
  \[
    \cF = \big\{ f_\by \in \{0,1\}^{\cX_T}: ~ \by \in \cY \big\}.
  \]
  Fix such a pair $(T,\cF)$. For each $\by \in \cY$ let
  \[
		P_{\by}\big((x, y)\big) = \sum_{\bu \in \left(\{0,1\}^{d}\right)^*} (d-1)d^{-\indexFunc{\bu}-1} \sum_{i=1}^d \1\left(x = \bx_{\bu}^i ~ \land ~ y = \customFunctionRound{f_\by}{\bx_{\bu}^i}\right).
	\] 
  In words, $P_{\by}$ corresponds to the following sampling procedure:
	\begin{enumerate}
		\item{\label{item:sampling-proc-1}
			Sample an index $k \in \bbN$ such that $\forall s \in \bbN: ~ \PP{k = s} = (d-1)d^{-s}$.\footnote{Recall that for a geometric series, $\sum_{s = 1}^\infty d^{-s} = \frac{1}{d-1}$ when $d > 1$, and therefore $\sum_{s = 1}^\infty \PP{k = s} = \sum_{s = 1}^\infty(d-1)d^{-s} = 1$.}
    }
    \item{
      Let $\bu \in \left(\{0,1\}^{d}\right)^*$ be the $k$-th string in the lexicographical ordering of $\left(\{0,1\}^{d}\right)^*$.
		}
		\item{\label{item:sampling-proc-2}
			Sample $j \in [d]$ independently and uniformly at random.
		}
		\item{\label{item:sampling-proc-3}
			Output $\left(\bx_{\bu}^j, \customFunctionRound{f_\by}{\bx_{\bu}^j}\right)$.
		}
	\end{enumerate}

  Note that the marginal distribution of $P_\by$ on $\cX$ (the distribution of $\bx_\bu^i$) is the same for all $\by \in \cY$; this is the marginal distribution $\cD_\cX$ mentioned in the statement.

  To see that $P_\by$ is realizable, note that for every $\varepsilon > 0$ there exists $k_\varepsilon \in \bbN$ such that in Step~\ref{item:sampling-proc-1} of the sampling procedure, $\PP{k > k_\varepsilon} \leq \varepsilon$. $f_\by$ is finitely-realizable by $\cH$ (\cref{item:finitely-realizable} in \cref{claim:properties-of-branches}), so in particular there exists $h_\varepsilon \in \cH$ that is consistent with $Z_\varepsilon = \bigl\{(\bx_{\bu}^j, f_\by(\bx_{\bu}^j)): ~ \indexFunc{\bu} \leq k_\varepsilon ~ \land ~ j \in [d]\bigr\}$. Hence, $\loss{P_\by}{h_\varepsilon} \leq \PPP{(x,y) \sim P_\by}{(x,y) \notin Z_\varepsilon} \leq \PP{k > k_\varepsilon} \leq \varepsilon$.

  For a fixed algorithm $\hh$ and for each $n \in \bbN$, consider the following experiment:
  \begin{itemize}
    \item{
      A value $\by \in \cY$ is sampled from the uniform distribution $\uniform{\cY}$, namely each bit in $\by$ is chosen independently and uniformly at random. 
    }
    \item{
      An i.i.d.\ training set $S = \big((X_1,Y_1,K_1),(X_2,Y_2,K_2), \dots, (X_n,Y_n,K_n)\big) \sim P_\by^n$ is generated according to the sampling procedure of Steps \ref{item:sampling-proc-1} to \ref{item:sampling-proc-3}, where for each $i \in [n]$, $K_i \in \bbN$ is the index selected at Step \ref{item:sampling-proc-1}, and $(X_i,Y_i)$ is the output at Step \ref{item:sampling-proc-3}. 
    }
    \item{
      An additional test sample $(X,Y,K) \sim P_\by$ is generated in the same manner.
    }
    \item{
      A randomness value $\rho$ is sampled for the algorithm $\hh$, and then $\hh$ is executed with training set $S$ and randomness $\rho$ and produces a hypothesis $\hh_S$.  
    }
    \item{
      $\hh_S$ is used to predict a label $\hh_S(X)$ for $X$. 
    }
  \end{itemize}
  This experiment defines a joint distribution 
  \begin{equation}\label{eq:lower-bound-experiment}
    (\by,S,X,Y,K,\rho)  
  \end{equation}
  that is used throughout the remainder of the proof.

  For any $\kappa \in \bbN$, let $\lowerBoundEvent(\kappa)$ denote the event in which the following conditions hold:
  \begin{itemize}
    \item $K = \kappa \geq \max\{K_1,\dots,K_n\}$.
    \item $|\{i \in [n]: K_i = \kappa\}|< d/2$.
    \item $X \notin \{X_i: ~ i \in [n]\}$.
  \end{itemize}

  We make two observations concerning $G(\kappa)$. The first observation is that 
  \begin{equation}\label{eq:observation-1}
    \PP{\lowerBoundEvent(\kappa)} \geq (d-1)d^{-\kappa}/4
  \end{equation}
  when $n = n_\kappa = \left\lfloor\frac{d^{\kappa + 1}}{8(d-1)}\right\rfloor$. To see this, let
  \begin{equation*}
    C_{=\kappa} = \left|\left\{i \in [n_\kappa]: ~ K_i = \kappa\right\}\right|, \quad C_{>\kappa} = \left|\left\{i \in [n_\kappa]: ~ K_i > \kappa\right\}\right|.
  \end{equation*}
  Then
  \begin{align*}
    \EE{C_{=\kappa}} = n_\kappa \cdot (d-1)d^{-\kappa} \leq \frac{d^{\kappa + 1}}{8(d-1)} \cdot (d-1)d^{-\kappa} = \frac{d}{8},
  \end{align*}
  and 
  \begin{align*}
    \EE{C_{>\kappa}} &= n_\kappa \cdot \sum_{s = \kappa+1}^\infty(d-1)d^{-s} \\
    &\leq \frac{d^{\kappa + 1}}{8(d-1)} \cdot 2(d-1)d^{-\kappa-1} = \frac{1}{4}.
  \end{align*}
  By Markov's inequality,
  \begin{align*}
    \PP{C_{=\kappa} \geq \frac{d}{2}} \leq \frac{1}{4}, \quad \text{and} \quad \PP{C_{>\kappa} \geq 1} \leq \frac{1}{4}.
  \end{align*}
  By a union bound, 
  \begin{align}\label{eq:bound-on-Ckappas}
    \PP{C_{=\kappa} < \frac{d}{2} ~ \land ~ C_{>\kappa} = 0} \geq \frac{1}{2}.
  \end{align}
  Hence, for any $\bleta \in \cY$,
  \begin{flalign*}
    & \PPP{\by,S,X,Y,K}{\lowerBoundEvent(\kappa) ~ | ~ \by = \bleta} 
    & \\ &
    \quad = \PP{K = \kappa ~ | ~ \by = \bleta} \cdot \PP{C_{=\kappa} < \frac{d}{2} ~ \land ~ C_{>\kappa} = 0 ~ | ~ \by = \bleta} 
    & \\ &
    \quad \quad \: \cdot \PP{X \notin \{X_i: ~ i \in [n_\kappa]\} ~ \Big| ~ C_{=\kappa} < \frac{d}{2} ~ \land ~ C_{>\kappa} = 0 ~ \land ~ K = \kappa ~ \land ~ \by = \bleta} 
    & \\ &
    \quad \geq (d-1)d^{-\kappa} \cdot \frac{1}{2}
    & \\ &
    \quad \quad \: \cdot \PP{X \notin \{X_i: ~ i \in [n_\kappa]\} ~ \Big| ~ C_{=\kappa} < \frac{d}{2} ~ \land ~ C_{>\kappa} = 0 ~ \land ~ K = \kappa ~ \land ~ \by = \bleta} \\
    \tag{By \cref{eq:bound-on-Ckappas}, $(C_{=\kappa},C_{>\kappa})\bot\by$}
    & \\ &
    \quad \geq (d-1)d^{-\kappa} \cdot \frac{1}{2} \cdot \frac{1}{2}.
    &
  \end{flalign*}
  For the last inequality, recall that the elements in $T$ are unique (\cref{item:unique-elements} in \cref{claim:properties-of-branches}). 
  Consequently, for every $i \in [n_\kappa]$, if $K_i < \kappa = K$ then $X_i \neq X$. The conditions $C_{= \kappa} < d/2$ and $K = \kappa$, and the sampling of $j \sim \uniform{[d]}$ in Step~\ref{item:sampling-proc-2} imply that with probability at least $1/2$, $X \notin \{X_i: ~ i \in [n_\kappa] ~ \land ~ K_i = \kappa\}$. This establishes \cref{eq:observation-1}, which is our first observation about $G(\kappa)$. 
  
  Our second observation is that for any $\kappa$ corresponding to a node on the branch $\by$, if $\lowerBoundEvent(\kappa)$ occurs then $\hh$ makes an incorrect prediction with probability $1/2$.

  Formally, for any $t \in \bbN$, let $\kappa_{\by,t} = \indexFunc{\by_{< t}}$, where $\by = (\by_1,\by_2,\dots)$ and $\by_{< t} = (\by_1,\by_2,\dots,\by_{t-1})$. In words, $\kappa_{\by,t}$ is the index in the lexicographic ordering of $(\{0,1\}^d)^*$ corresponding to the $t$-th node in the branch $\by$. Let $n_{\by,t} = n_{\kappa_{\by,t}}$. The second observation states that for any $t \in \bbN$,
  \begin{flalign}\label{eq:observation-2}
    &
    \EEE{\by \sim \uniform{\cY}}{\PPP{S \sim P_\by^{n_{\by,t}},(X,Y,K) \sim P_{\by},\rho}{\hh_S(X) \neq Y ~ | ~ \lowerBoundEvent(\kappa_{\by,t})}}
    & \nonumber \\ &
    \qquad = \PPP{\by \sim \uniform{\cY},S \sim P_\by^{n_{\by,t}},(X,Y,K) \sim P_{\by},\rho}{\hh_S(X) \neq Y ~ | ~ \lowerBoundEvent(\kappa_{\by,t})} = \frac{1}{2}.
    &
  \end{flalign}
  This probability pertains to the special case of the experiment of \cref{eq:lower-bound-experiment} in which the number $n$ of samples in $S$ depends on $\by$, satisfying $n = n_{\by,t}$. 
  It is a conditional probability given that $\lowerBoundEvent(\kappa_{\by,t})$ occurred, where $\lowerBoundEvent(\kappa_{\by,t})$ is an event involving $(\by,X,X_1,\dots,X_n,K,K_1,\dots,K_n)$.
  To establish \cref{eq:observation-2}, it suffices to show that for any $t \in \bbN$,
  \begin{equation}\label{eq:Y-is-bernoulli-given-XiYiX}
    \PPP{\by \sim \uniform{\cY},S \sim P_\by^{n_{\by,t}},(X,Y,K) \sim P_{\by},\rho}{Y = 1 ~ | ~ X, \{X_i,Y_i\}_{i \in [n_{\by,t}]}, G(\kappa_{\by,t})} = \frac{1}{2},
  \end{equation}
  because the prediction $\hh_S(X)$ depends only on $(X, \{X_i,Y_i\}_{i \in [n_{\by,t}]},\rho)$.
  Roughly, \cref{eq:Y-is-bernoulli-given-XiYiX} follows from the indifference of $\{f_\by\}_{\by \in \cY}$ (\cref{item:indifference-property-branches} in \cref{claim:properties-of-branches}), which states that if $X$ is a member of the $K$-th node in the tree $T$, then for any $X_i$ with $K_i < K$ there exists a bit $b \in \{0,1\}$ such that for all branches $\by \in \cY$ that contain node $K$, $f_\by(X_i) = b$. In particular, $Y = f_\by(X)$ is a uniformly random bit independent of $\{X_i,Y_i = f_\by(X_i)\}_{i \in [n_{\by,t}]}\cup\{X\}$ given $G(\kappa_{\by,t})$.

  To flesh out the argument for \cref{{eq:Y-is-bernoulli-given-XiYiX}} in further detail, fix $\kappa \in \bbN$, $(\kappa_1,\dots,\kappa_{n_{\kappa}}) \allowbreak \in \bbN^{n_{\kappa}}$, $(\xi,\xi_1,\dots,\xi_{n_{\kappa}}) \in \cX^{n_{\kappa}+1}$, and $(\eta_1,\dots,\eta_{n_{\kappa}}) \in \{0,1\}^{n_{\kappa}}$. Consider the following conditional probability of $Y$ for a fixed $t \in \bbN$, assuming the event being conditioned upon has a positive probability. 
  \begin{flalign*}
    & \PP{Y = 1 ~ \Bigg| ~ \left.
      \begin{array}{c}
        \kappa_{\by,t} = \kappa \\ 
        \forall i \in  [n_{\by,t}]: ~ X_i = \xi_i ~ \land ~ Y_i = \eta_i ~ \land ~ K_i = \kappa_i \\
        K = \kappa_{\by,t} \geq \max\{K_i: ~ i \in [n_{\by,t}]\} \\
        X = \xi\notin \{X_i: ~ i \in [n_{\by,t}]\}
      \end{array}
    \right.}
    & \\ &
    \qquad =
    \PP{f_\by(X) = 1 ~ \Bigg| ~ \left.
      \begin{array}{c}
        \kappa_{\by,t} = \kappa \\ 
        \forall i \in  [n_{\by,t}]: ~ X_i = \xi_i ~ \land ~ f_\by(X_i) = \eta_i ~ \land ~ K_i = \kappa_i \\
        K = \kappa_{\by,t} \geq \max\{K_i: ~ i \in [n_{\by,t}]\} \\
        X = \xi\notin \{X_i: ~ i \in [n_{\by,t}]\}
      \end{array}
    \right.}
    \\
    \tag{Choice of $Y$ and $Y_i$}
    & \\ &
    \qquad =
    \PP{f_\by(X) = 1 ~ \Bigg| ~ \left.
      \begin{array}{c}
        \kappa_{\by,t} = \kappa \\ 
        \forall i \in  [n_{\by,t}]: ~ X_i = \xi_i ~ \land ~ K_i = \kappa_i \\
        K = \kappa_{\by,t} \geq \max\{K_i: ~ i \in [n_{\by,t}]\} \\
        X = \xi\notin \{X_i: ~ i \in [n_{\by,t}]\}
      \end{array}
    \right.}
    & \\ &
    \tag{Indifference of $\{f_\by\}_{\by \in \cY}$ -- \cref{item:indifference-property-branches} in \cref{claim:properties-of-branches}} 
    & \\ &
    \qquad =
    \PP{\by_t^j = 1 ~ \Bigg| ~ \left.
      \begin{array}{c}
        \kappa_{\by,t} = \kappa \\ 
        \forall i \in  [n_{\by,t}]: ~ X_i = \xi_i ~ \land ~ K_i = \kappa_i \\
        K = \kappa_{\by,t} \geq \max\{K_i: ~ i \in [n_{\by,t}]\} \\
        X = \xi\notin \{X_i: ~ i \in [n_{\by,t}]\}
      \end{array}
    \right.}
    = \frac{1}{2},
    &
  \end{flalign*}
  where $j$ is the index of $X$ in the $K$-th node in the tree. In the last line we have used the fact that $K = \kappa_{\by,t}$ implies that $X$ is on the branch corresponding to $\by$, and the final equality holds because $\by_t$ is a vector of uniformly random bits chosen independently of $\{X_i,K_i\}_{i \in [n_{\by,t}]} \cup \{X,K,\kappa_{\by,t}\}$ (note that $\kappa_{\by,t}$ and $n_{\by,t}$ are fully determined by $t$ and $\by_{<t}$). 
  This establishes \cref{eq:observation-2}, our second observation.

  The first observation is used as follows. For every $\by \in \cY$,
  \begin{flalign}
    &\limsup_{n \rightarrow \infty}n\cdot\EEE{\rho,S \sim P_\by^n}{\loss{P_{\by}}{\hh_{S}}}
    & \nonumber\\ &
    \quad \geq \limsup_{t \rightarrow \infty}n_{\by,t}\cdot\EEE{\rho,S \sim P_\by^{n_{\by,t}}}{\loss{P_{\by}}{\hh_{S}}}
    \tag{If $b_j$ is a subsequence of $a_j$ then $\limsup a_j \geq \limsup b_j$}
    & \nonumber\\ &
    \quad = \limsup_{t \rightarrow \infty} n_{\by,t}\cdot\PPP{\rho,S \sim P_\by^{n_{\by,t}},(X,Y,K) \sim P_{\by},\rho}{\hh_{S}(X) \neq Y}
    & \nonumber\\ &
    \quad \geq \limsup_{t \rightarrow \infty}n_{\by,t}\cdot\PPP{\rho,S \sim P_\by^{n_{\by,t}},(X,Y,K) \sim P_{\by}}{\left(\hh_{S}(X) \neq Y\right) ~ \land ~ \lowerBoundEvent(\kappa_{\by,t})}
    & \nonumber\\ &
    \quad = \limsup_{t \rightarrow \infty}n_{\by,t}\cdot\PP{\lowerBoundEvent(\kappa_{\by,t})}\cdot\PP{\hh_S(X) \neq Y ~ | ~ \lowerBoundEvent(\kappa_{\by,t})}
    & \nonumber\\ &
    \quad \geq \limsup_{t \rightarrow \infty}\frac{d^{\kappa_{\by,t} + 1}}{9(d-1)}\cdot \frac{(d-1)d^{-\kappa_{\by,t}}}{4}\cdot\PP{\hh_S(X) \neq Y ~ | ~ \lowerBoundEvent(\kappa_{\by,t})}
    \tag{By \cref{eq:observation-1} and choice of $n_{\by,t}$}
    & \nonumber\\ &
    \quad = \limsup_{t \rightarrow \infty}\frac{d}{36}\cdot\PPP{\rho,S \sim P_\by^{n_{\by,t}},(X,Y,K) \sim P_{\by}}{\hh_S(X) \neq Y ~ | ~ \lowerBoundEvent(\kappa_{\by,t})}.
    \label{eq:lower-bound-on-n-prob}
  \end{flalign}
  To complete the proof we use our second observation and Fatou's lemma as follows.
  \begin{flalign*}
    &\EEE{\by \sim \uniform{\cY}}{\limsup_{n \rightarrow \infty}n\cdot\EEE{\rho,S \sim P_\by^n}{\loss{P_{\by}}{\hh_{S}}}} 
    & \\ &
    \quad \geq \frac{d}{36}\cdot\EEE{\by \sim \uniform{\cY}}{\limsup_{t \rightarrow \infty}\PPP{\rho,S \sim P_\by^{n_{\by,t}},(X,Y,K) \sim P_{\by}}{\hh_S(X) \neq Y ~ | ~ \lowerBoundEvent(\kappa_{\by,t})}}
    \tag{By \cref{eq:lower-bound-on-n-prob}}
    & \\ &
    \quad \geq \frac{d}{36}\cdot\limsup_{t \rightarrow \infty}\EEE{\by \sim \uniform{\cY}}{\PPP{\rho,S \sim P_\by^{n_{\by,t}},(X,Y,K) \sim P_{\by}}{\hh_S(X) \neq Y ~ | ~ \lowerBoundEvent(\kappa_{\by,t})}}
    \tag{Fatou's lemma (\cref{lemma:reverse-fatou}), $\PP{\cdot} \leq 1$}
    & \\ &
    \quad = \frac{d}{36}\cdot\frac{1}{2} = \frac{d}{72}.
    \tag{By \cref{eq:observation-2}}
    &
  \end{flalign*}
  This implies that there exists $\by \in \cY$ such that 
  \[
    \limsup_{n \rightarrow \infty}n\cdot\EEE{S \sim P_\by^n}{\loss{P_{\by}}{\hh_{S}}} \geq \frac{d}{72}.
  \]
  By the definition of $\limsup$, the inequality
  \[
    \EEE{S \sim P_\by^n}{\loss{P_{\by}}{\hh_{S}}} \geq \frac{d}{73\cdot n}
  \]
  holds for infinitely many values of $n \in \bbN$, as desired.
\end{proof}

\section{Result for Half-Spaces}\label{section:half-spaces}

\begin{notation}
	Let $d \in \bbN$. We write $\bbS^{d-1} = \{x \in \bbR^d: ~ \|x\|_2 = 1\}$ to denote the unit sphere in $\bbR^d$.
\end{notation}

\begin{definition}
	Let $d \in \bbN$. For any $\bw \in \bbS^{d-1}$, let $h_{\bw}: ~ \bbR^d \rightarrow \{0,1\}$ be the half-space given by $h_\bw(\bx) = \1\left(\left\langle \bw,\bx \right\rangle > 0\right)$. The \ul{class of homogeneous half-spaces in $\bbR^d$} is $\cH_d = \left\{h_{\bw}: ~ \bw \in \bbS^{d-1}\right\}$.
\end{definition}

\begin{definition}
	Let $d \in \bbN$, let $H \subseteq \bbS^{d-1}$ be a set. We say that a set of points $\{x_1,\dots,x_m\} \subseteq \bbR^d$ is \ul{openly shattered by $H$} if for every vector $\by = (y_1,\dots,y_m) \in \{0,1\}^m$, there exists an open set $W_\by \subseteq H$ such that
	\begin{equation}\label{eq:open-shatter}
		\forall \bw \in W_\by ~ \forall i \in [m]: ~ h_{\bw}(x_i) = y_i.
	\end{equation}
\end{definition}

\begin{lemma}\label{lemma:openly-shattered}
	Let $d \in \bbN$, and let $H \subseteq \bbS^{d-1}$ be an open set. Then there exists a set $X \subseteq \bbS^{d-1}$ such that $|X| = d-1$ and $X$ is openly shattered by $H$.
\end{lemma}

\begin{proof}
	Fix a point $x_0$ in the interior of the $H$. Let $x_1,\dots,x_{d-1} \in \bbS^{d-1}$ be points such that $x_0,x_1,\dots,x_{d-1}$ is an orthonormal basis of $\bbR^d$. 
	
	For each $\by = (y_1,\dots,y_{d-1}) \in \{0,1\}^{d-1}$, let 
	\[
		w_\by' = x_0 + \varepsilon \cdot \sum_{i \in [d-1]} \mathrm{sign}(y_i -\sfrac{1}{2}) \cdot x_i
	\]
	be a point with $\varepsilon > 0$ small enough such that $w_\by$ is in the interior of $H$, where $w_\by$ is the projection of $w_\by'$ onto $\bbS^{d-1}$. From the orthogonality of $\{x_0,\dots,x_{d-1}\}$, 
	\begin{equation*}
		\forall i \in [d-1]: ~ h_{w_\by}(x_i) = y_i.
	\end{equation*}
	
	For each $i \in [d-1]$ and $y \in \{0,1\}$, let $Q_{i,y} \subseteq \bbS^{d-1}$ be the set of $\bw$ such that $h_\bw(x_i) = y$. Because we use open half-spaces, $Q_{i,y}$ is open. Observe that for each $\by \in \{0,1\}^{d-1}$,
	\[
		W_\by = H \cap \bigcap_{i \in [d-1]} Q_{i,y_i}
	\]
	is open (as a finite intersection of open sets), and is non-empty because it contains $w_\by$.
\end{proof}

\begin{lemma}
	Let $d \in \bbN$. Then $\VCL{\cH_d} \geq d-1$.
\end{lemma}

\begin{proof}
	We recursively construct an infinite $(d-1)$-VCL tree that is shattered by $\cH_d$. Let $H_\lambda = \bbS^{d-1}$. For every $s \in 0,1,2,...$ do the following. For every $\bu \in \{0,1\}^{ds}$, note that $H_\bu \subseteq \bbS^{d-1}$ is open. Therefore, by \cref{lemma:openly-shattered}, there exists $\bx_\bu = (x_\bu^1,\dots,x_\bu^{d-1})\subseteq \bbS^{d-1}$ of cardinality $d-1$ that is openly shattered by $H_\bu$. Namely, for each $\by \in \{0,1\}^d$ there exists an open set $W_\by \subseteq H_\bu$ such that \cref{eq:open-shatter} holds (for $x_i = x_\bu^i$ and $m=d$). For each $\by \in \{0,1\}^d$, define $H_{\bu \circ \by} = W_\by$.
	
	We claim that $T = \{\bx_\bu: ~ \bu \in \left(\{0,1\}^{d}\right)^*\}$ is a $(d-1)$-VCL tree that is shattered by $\cH_d$. Indeed, fix $t \in \bbN$ and $\by \in \{0,1\}^{td}$. Let $\bw \in H_\bu$. Then the choice of $\bx_\bu$ and $H_\bu$ implies that
	\[
		\forall s \in [t] \: \forall j \in [d]: ~ h_\bw(x_{\by_{\leq s-1}}^j) = y_{s}^j,
	\]
	as desired.
\end{proof}

\section{Directions for Future Work}

We have shown a characterization of fine-grained learning rates in the instance specific setting. Directions for future work include characterizing the precise parameters $C,c \geq 0$ in \cref{eq:general-form-of-learning-curve}, and obtaining an optimal gap factor (or equivalently, optimal parameters $\alpha,\beta \geq 0$ in \cref{theorem:fine-grained}).

Like the results of \cite{DBLP:conf/stoc/BousquetHMHY21}, our results describe the \emph{asymptotic} rate at which learning curves decay -- but the results are silent as to the properties of learning curves for any finite number of samples. Devising a theory of learning curves that explains both asymptotic and non-asymptotic behavior in a unified way would be valuable.

Our result on semi-supervised learning (\cref{item:main-result-ssl} in \cref{section:main-results}, Main Results) suggests that unlabeled data is not helpful in the setting of distribution-dependent learning curves. However, there are good reasons to believe that unlabeled data is helpful for learning in some real-world scenarios. We wonder how this tension could be resolved.

\subsection*{Acknowledgments}

{\footnotesize
  SM is a Robert J.\ Shillman fellow. His research is supported in part
  by the Israel Science Foundation (grant No.\ 1225/20), by a grant from the
  United States -- Israel Binational Science Foundation (BSF), by an Azrieli
  Faculty Fellowship, by Israel PBC-VATAT, and by the Technion Center for
  Machine Learning and Intelligent Systems (MLIS).

  JS would like to thank Robert D.\ Kleinberg and SM for their wise mentorship and wholehearted support while hosting him at Cornell University and at the Technion, respectively. Part of this work was conducted while visiting those institutions. This work was supported in part by a grant from the Simons Foundation (\#733786, UCB), by the Defense Advanced Research Projects Agency (DARPA) under award \#HR00112020021 (sub-award \#4500003487), and by SM's Azrieli Faculty Fellowship. Any opinions, findings and conclusions or recommendations expressed in this material are those of the author(s) and do not necessarily reflect the views of the Simons Foundation, DARPA, or the Azrieli Foundation.

  IT would like to thank Konstantin Golubev for useful discussions.
}

\setcitestyle{numbers}
\bibliographystyle{plainnat}
\bibliography{paper,learning}

\begin{thebibliography}{16}
\providecommand{\natexlab}[1]{#1}
\providecommand{\url}[1]{\texttt{#1}}
\expandafter\ifx\csname urlstyle\endcsname\relax
  \providecommand{\doi}[1]{doi: #1}\else
  \providecommand{\doi}{doi: \begingroup \urlstyle{rm}\Url}\fi

\bibitem[wik(2022)]{wiki:FatousLemma}
Reverse {F}atou's {L}emma.
\newblock \emph{{P}roof{W}iki}, August 2022.
\newblock URL \url{https://proofwiki.org/wiki/Reverse_Fatou%27s_Lemma}.

\bibitem[Antos and Lugosi(1998{\natexlab{a}})]{antos:98}
A.~Antos and G.~Lugosi.
\newblock Strong minimax lower bounds for learning.
\newblock \emph{Machine Learning}, 30:\penalty0 31--56, 1998{\natexlab{a}}.

\bibitem[Antos and Lugosi(1996)]{DBLP:conf/colt/AntosL96}
Andr{\'{a}}s Antos and G{\'{a}}bor Lugosi.
\newblock Strong minimax lower bounds for learning.
\newblock In Avrim Blum and Michael~J. Kearns, editors, \emph{Proceedings of
  the Ninth Annual Conference on Computational Learning Theory, {COLT} 1996,
  Desenzano del Garda, Italy, June 28-July 1, 1996}, pages 303--309. {ACM},
  1996.
\newblock \doi{10.1145/238061.238160}.
\newblock URL \url{https://doi.org/10.1145/238061.238160}.

\bibitem[Antos and Lugosi(1998{\natexlab{b}})]{DBLP:journals/ml/AntosL98}
Andr{\'{a}}s Antos and G{\'{a}}bor Lugosi.
\newblock Strong minimax lower bounds for learning.
\newblock \emph{Mach. Learn.}, 30\penalty0 (1):\penalty0 31--56,
  1998{\natexlab{b}}.
\newblock \doi{10.1023/A:1007454427662}.
\newblock URL \url{https://doi.org/10.1023/A:1007454427662}.

\bibitem[Ben{-}David and Blais(2020)]{DBLP:conf/focs/Ben-DavidB20a}
Shalev Ben{-}David and Eric Blais.
\newblock A new minimax theorem for randomized algorithms (extended abstract).
\newblock In Sandy Irani, editor, \emph{61st {IEEE} Annual Symposium on
  Foundations of Computer Science, {FOCS} 2020, Durham, NC, USA, November
  16-19, 2020}, pages 403--411. {IEEE}, 2020.
\newblock \doi{10.1109/FOCS46700.2020.00045}.
\newblock URL \url{https://doi.org/10.1109/FOCS46700.2020.00045}.

\bibitem[Bousquet et~al.(2021)Bousquet, Hanneke, Moran, van Handel, and
  Yehudayoff]{DBLP:conf/stoc/BousquetHMHY21}
Olivier Bousquet, Steve Hanneke, Shay Moran, Ramon van Handel, and Amir
  Yehudayoff.
\newblock A theory of universal learning.
\newblock In Samir Khuller and Virginia~Vassilevska Williams, editors,
  \emph{{STOC} '21: 53rd Annual {ACM} {SIGACT} Symposium on Theory of
  Computing, Virtual Event, Italy, June 21-25, 2021}, pages 532--541. {ACM},
  2021.
\newblock \doi{10.1145/3406325.3451087}.
\newblock URL \url{https://doi.org/10.1145/3406325.3451087}.

\bibitem[Browder(1996)]{browder1996}
Andrew Browder.
\newblock \emph{Mathematical analysis: an introduction}.
\newblock Springer Science \& Business Media, 1996.

\bibitem[Devroye et~al.(1996)Devroye, Gy\"{o}rfi, and Lugosi]{devroye:96}
L.~Devroye, L.~Gy\"{o}rfi, and G.~Lugosi.
\newblock \emph{A Probabilistic Theory of Pattern Recognition}.
\newblock Springer-Verlag New York, Inc., 1996.

\bibitem[Gale and Stewart(1953)]{GS53}
D.~Gale and F.~M. Stewart.
\newblock Infinite games with perfect information.
\newblock In \emph{Contributions to the theory of games, vol. 2}, Annals of
  Mathematics Studies, no. 28, pages 245--266. Princeton University Press,
  Princeton, N. J., 1953.

\bibitem[Haussler et~al.(1994)Haussler, Littlestone, and Warmuth]{haussler:94}
D.~Haussler, N.~Littlestone, and M.~Warmuth.
\newblock Predicting $\{0,1\}$-functions on randomly drawn points.
\newblock \emph{Information and Computation}, 115\penalty0 (2):\penalty0
  248--292, 1994.

\bibitem[Schuurmans(1997)]{schuurmans:97}
D.~Schuurmans.
\newblock Characterizing rational versus exponential learning curves.
\newblock \emph{Journal of Computer and System Sciences}, 55\penalty0
  (1):\penalty0 140--160, 1997.

\bibitem[Shalev{-}Shwartz and Ben{-}David(2014)]{DBLP:books/daglib/0033642}
Shai Shalev{-}Shwartz and Shai Ben{-}David.
\newblock \emph{Understanding Machine Learning - From Theory to Algorithms}.
\newblock Cambridge University Press, 2014.
\newblock ISBN 978-1-10-705713-5.
\newblock URL \url{https://doi.org/10.1017/CBO9781107298019}.

\bibitem[Stone(1977)]{stone:77}
C.~J. Stone.
\newblock Consistent nonparametric regression.
\newblock \emph{The Annals of Statistics}, pages 595--620, 1977.

\bibitem[Valiant(1984)]{DBLP:journals/cacm/Valiant84}
Leslie~G. Valiant.
\newblock A theory of the learnable.
\newblock \emph{Commun. {ACM}}, 27\penalty0 (11):\penalty0 1134--1142, 1984.
\newblock \doi{10.1145/1968.1972}.
\newblock URL \url{https://doi.org/10.1145/1968.1972}.

\bibitem[van Handel(2013)]{van-handel:13}
R.~van Handel.
\newblock The universal {G}livenko-{C}antelli property.
\newblock \emph{Probability and Related Fields}, 155:\penalty0 911--934, 2013.

\bibitem[Vapnik and Chervonenkis(1968)]{VapChe68}
V.~N. Vapnik and A.~Ya. Chervonenkis.
\newblock The uniform convergence of frequencies of the appearance of events to
  their probabilities.
\newblock \emph{Dokl. Akad. Nauk {SSSR}}, 181\penalty0 (4):\penalty0 781--783,
  1968.
\newblock URL \url{http://www.ams.org/mathscinet-getitem?mr=0231431}.

\end{thebibliography}




\end{document}